\documentclass[journal]{IEEEtran}

\usepackage{packages/algorithm}
\usepackage{packages/algorithmic}

\ifCLASSOPTIONcompsoc
\usepackage[caption=false,font=normalsize,labelfont=sf,textfont=sf]{subfig}
\else
\usepackage[caption=false,font=footnotesize]{subfig}
\fi

\usepackage{graphics} 
\usepackage{amsmath} 
\usepackage{mathtools} 
\usepackage[utf8]{inputenc} 
\usepackage[english]{babel}
\usepackage{multirow}

\usepackage{helvet}         
\usepackage{courier}        
\usepackage{type1cm}        
\usepackage{makeidx}         
\usepackage{graphicx}        
\usepackage{multicol}        
\usepackage[bottom]{footmisc}
\usepackage{subfig}  

\usepackage{tikz}
\usetikzlibrary{positioning}
\usetikzlibrary{shapes,snakes}
\usepackage{bm} 
\usepackage{float} 
\usepackage{color}

\usepackage{caption}
\usepackage{amsmath,amssymb} 
\usepackage{graphicx}
\usepackage{todonotes}

\usepackage{empheq} 
\usepackage{xcolor}

\usepackage[backend=biber, style=ieee, doi=false,isbn=false,url=false,eprint=false]{biblatex}
\bibliography{libraries/rpng,libraries/xins,libraries/extra}

\usepackage{upgreek,bm}
\usepackage{amsmath, amssymb, amsfonts}
\usepackage{graphicx}
\usepackage{todonotes}
\usepackage{float} 
\usepackage{xcolor} 
\usepackage{forest} 
\usepackage{listings} 
\usepackage{pdflscape} 
\usepackage{booktabs} 
\usepackage{adjustbox} 
\usepackage{caption}

\newcommand\scalemath[2]{\scalebox{#1}{\mbox{\ensuremath{\displaystyle #2}}}}

\usepackage{array}
\newcommand{\PreserveBackslash}[1]{\let\temp=\\#1\let\\=\temp}
\newcolumntype{C}[1]{>{\PreserveBackslash\centering}m{#1}}
\newcolumntype{R}[1]{>{\PreserveBackslash\raggedleft}p{#1}}
\newcolumntype{L}[1]{>{\PreserveBackslash\raggedright}p{#1}}

\allowdisplaybreaks
 \renewcommand{\baselinestretch}{0.97}
\begin{document}

\title{MIMC-VINS: A Versatile and Resilient Multi-IMU Multi-Camera Visual-Inertial Navigation System}

\author{Kevin Eckenhoff,~\IEEEmembership{Student Member,~IEEE,} Patrick Geneva,~\IEEEmembership{Student Member,~IEEE,} and Guoquan~Huang,~\IEEEmembership{Member,~IEEE}
\thanks{The authors are with the Robot Perception and Navigation Group (RPNG), University of Delaware, Newark, DE 19716, USA.
Email: {\tt \{keck,pgeneva,ghuang\}@udel.edu}.}
}

\maketitle

\begin{abstract}
As cameras and inertial sensors are becoming ubiquitous in mobile devices and robots, 
it holds great potential to design visual-inertial navigation systems (VINS) for efficient versatile 3D motion tracking
which utilize any (multiple) available cameras and inertial measurement units (IMUs)
and are resilient to sensor failures or measurement depletion.
To this end, rather than the standard VINS paradigm using a minimal sensing suite of a single camera and IMU, 
in this paper we design a real-time consistent multi-IMU multi-camera (MIMC)-VINS estimator 
that is able to seamlessly fuse multi-modal information from an arbitrary number of uncalibrated cameras and IMUs.
Within an efficient multi-state constraint Kalman filter (MSCKF) framework,
the proposed MIMC-VINS algorithm optimally fuses asynchronous measurements from all sensors,
while providing smooth, uninterrupted, and accurate 3D motion tracking even if some sensors fail.
The key idea of the proposed MIMC-VINS is to perform high-order on-manifold state interpolation to efficiently process all available visual measurements without increasing the computational burden due to estimating additional sensors' poses at asynchronous imaging times. In order to fuse the information from multiple IMUs, we propagate a joint system consisting of all IMU states while enforcing rigid-body constraints between the IMUs during the filter update stage. Lastly, we estimate online both spatiotemporal extrinsic and visual intrinsic parameters to make our system robust to errors in prior sensor calibration.
The proposed system is extensively validated in both Monte-Carlo simulations and real-world experiments.
\end{abstract}


\section{Introduction} \label{sec::intro}

As cameras and inertial sensors are commonplace in today's mobile devices and autonomous vehicles,
developing visual-inertial navigation systems (VINS) for 3D motion tracking has been arguably at the center of recent simultaneous localization and mapping (SLAM) research efforts~\cite{Huang2019ICRA}.
However, most of the current VINS algorithms have focused on the case of minimal sensing where only a {\em single} camera and inertial measurement unit (IMU) is considered 
(e.g., see~\cite{Hesch2013TRO,Hesch2014IJRR,Li2013IJRR,Mourikis2007ICRA,Qin2018TRO}). While 3D motion tracking with minimal sensing capability is of interest, 
in practice, it is desirable to optimally and efficiently fuse \textit{all} information from {\em multiple} visual-inertial sensors  
to improve estimation robustness and accuracy~\cite{Paul2017ICRA}. In addition,
given the fact that nowadays these sensors have become small and affordable, in many scenarios it has become feasible and even necessary to utilize multiple sensors.
In particular, we note that most single-view systems are susceptible to loss of texture in a given viewing direction~\cite{Engel2018TPAMI}, and thus a single-camera system may suffer greatly from measurement depletion (i.e. no informative measurements are available). 
However, even with multiple cameras, conditions such as poor lighting may cause the estimator to rely solely on its IMU due to the lack of visual information. 
As such, developing visual-inertial systems that leverage multiple sensors (both cameras and IMUs) is of practical significance
and enabling resilience of such systems to sensor failures is crucial in practice.

Fusing information from multiple sensors which  collect local information requires knowing the spatial transformation (relative pose) between each sensor. Moreover, if the sensors are not hardware synchronized (i.e. they are not electronically triggered to collect data at the same time) then representing the state at each sensing time may become computationally infeasible.
Additionally, due to latency issues, data transfer times, or different independent sensor clocks, nontrivial time offsets between different sensor measurement timestamps present an additional barrier to accurate state estimation \cite{Li2014IJRRa}.
This can be addressed by manufacturing all sensing components as a single tightly-coupled time-synchronized unit, 
but may become prohibitively expensive for widespread applications.
Clearly, non-tightly-coupled sensor design has great impact to endow VINS with \textit{plug-and-play} functionality,
wherein different sensors can be freely added/removed without requiring hardware synchronization or exhaustive offline sensor calibration, if these issues are addressed.
Plug-and-play functionality significantly lowers the technology barriers for end users and thus promotes this emerging technology in many different application domains such as augmented reality and autonomous driving.

To achieve the aforementioned plug-and-play functionality, 
building upon our recent conference publications~\cite{Eckenhoff2019ICRAa,Eckenhoff2019ICRAb},
in this paper we design a versatile and resilient  multi-IMU multi-camera (MIMC)-VINS algorithm 
that can utilize an arbitrary number of uncalibrated and asynchronous cameras and IMUs.
Within an efficient multi-state constraints Kalman filter (MSCKF) framework~\cite{Mourikis2007ICRA}, 
the proposed MIMC-VINS estimator is able to fuse the information from all sensors
while providing smooth, uninterrupted, accurate 3D motion tracking even if some sensors fail.
The key idea of the proposed MIMC-VINS is to perform high-order on-manifold state interpolation to efficiently process all available visual measurements without increasing the computational burden due to estimating additional sensors' poses at asynchronous imaging times. In order to fuse the information from multiple IMUs, we propagate a joint system consisting of all IMU states while enforcing rigid-body constraints between the IMUs during the filter update stage. Additionally, we estimate online both spatiotemporal extrinsic and camera intrinsic parameters to make our system robust to errors in prior sensor calibration. Lastly, we enforce the well-known VINS observability constraints in computing Jacobians to improve estimation consistency.

In particular, the main contributions of this paper include:
\begin{itemize}

\item We develop a real-time, easy-to-use, versatile MIMC-VINS state estimator with online sensor calibration,
which can utilize an arbitrary number of uncalibrated asynchronized cameras and IMUs 
while performing online calibration of all sensing parameters including visual intrinsics and spatial/temporal extrinsics.
In particular, due to the growing computation required to process increased measurements provided by more sensors,
leveraging the lightweight MSCKF framework, the proposed MIMC-VINS estimator focuses on the seamless and efficient incorporation of multiple sensors.
\item We further advance the MIMC-VINS estimator by adapting the first-estimate Jacobian (FEJ) observability-based methodology~\cite{Huang2008ICRA} to improve consistency,
and by introducing high-order polynomial fitting for on-manifold interpolation to accurately fuse asynchronous sensor measurements at low computational cost.
\item 
The proposed MIMC-VINS is able to offer resilience to sensor failures 
and robustness to measurement depletion of single views (due to the lack of texture in a viewing direction) by utilizing redundant sensors.

\item We thoroughly validate the proposed MIMC-VINS 
using different number of visual and inertial sensors 
in both Monte-Carlo simulations and real-world experiments,
in terms of calibration convergence and estimation accuracy, consistency, and resilience to sensor failures.
\end{itemize}

The rest of the paper is structured as follows: 
After reviewing the related work in the next section, we provide the necessary background about single-camera single-IMU MSCKF-based VINS in Section~\ref{sec:vins}.
We present in detail the barebones multi-camera multi-IMU (MIMC)-VINS in Section~\ref{sec:mimc-vins},
which is future advanced to perform online sensor calibration and enable consistent resilient estimation
in Section~\ref{sec:mimc-ext}. 
The proposed MIMC-VINS is validated extensively in Sections~\ref{sec:sim} and \ref{sec:exp}.
Finally, we conclude the paper in Section~\ref{sec:concl} along with possible future research directions.

\section{Related Work}

In part due to the recent advancements of these two complementary sensing technologies, 
visual-inertial state estimation has attracted significant research attentions in recent years~\cite{Huang2019ICRA}. 
In this section, we briefly review the related VINS literature with multiple sensors.

\subsection{Multi-Camera Systems}

While monocular-VINS has been widely studied (e.g., see~\cite{Huang2019ICRA,Hesch2013TRO,Hesch2014IJRR,Huai2018IROS,Huang2014ICRA, Li2013IJRR,Mourikis2009TRO,Qin2018TRO,Geneva2020ICRA} and references therein),
one straightforward extended configuration over the monocular setting is to use a {\em stereo} camera,
wherein {\em two} cameras are mounted such that they observe the same spatial volume from the offset camera centers at the same imaging time.
Stereo vision enables 3D triangulation of features seen in the overlapping view without requiring motion of the sensor platform, thus allowing for the direct recovery of scale if the spatial transforms between cameras are known.
Motivated by this increase in robustness, Sun et al.~\cite{Sun2018RAL} developed the MSCKF-based stereo-VINS with the particular application to high-speed aerial vehicles.
Paul et al.~\cite{Paul2017ICRA} extended the inverse square-root version of the MSCKF (namely SR-ISWF) \cite{Wu2015RSS} to provide real-time VINS on mobile devices while allowing for a configuration of both stereo and binocular (non-overlapping) cameras, 
and showed that the inclusion of more cameras improves the estimation accuracy.

While stereo cameras provide robustness due to their ability to perform feature triangulation and scale recovery even without the IMU, they remain vulnerable to dynamic environmental motion and textureless regions in its given viewing direction.
More importantly, the requirement of an overlapping field of view and synchronous camera triggering may not easily extend to 
an {\em arbitrary} number of {\em plug-and-play} cameras -- 
which is a highly-desirable characteristic and could greatly promote the widespread deployment of VINS in practice.
Additionally, due to the enforcement of cross-image matching (for example matching features from the left to right stereo image), 
the process of visual tracking is coupled and cannot be directly parallelized to facilitate a large number of cameras.
For these reasons, in our prior work~\cite{Eckenhoff2019ICRAa},
we introduced a general multi-camera VINS algorithm, which can tightly fuse the visual information from an {\em arbitrary} number of \textit{non-overlapping}, {\em asynchronous} heterogeneous cameras and an IMU, 
so that our approach is robust to environmental conditions and single-camera failures while allowing for improved estimation performance.
Note that in our multi-camera VINS system, we do {\em not} perform any cross-image matching,
since we have non-overlapping images and instead allow each camera image stream to be processed independently and in parallel.
In this work we further generalize this system to multi-camera {\em and} multi-IMU scenarios.

Houben et al.~\cite{Houben2016IROS} extended the ORB-SLAM~\cite{Mur-Artal2015TRO} to a system of  multiple cameras with varying viewing directions and an IMU for UAVs within a graph-SLAM framework but assumed known sensor calibration and simultaneous triggering of all involved cameras.
Paul et al.~\cite{Paul2018CVPR} addressed the problem of increased computational burden in stereo-VINS and proposed an alternating stereo-VINS algorithm. 
In their system, the {\em two} cameras in a stereo pair were triggered in an alternating manner, preventing the need to process both images at the same time while still taking advantage of the offset camera centers provided by a stereo configuration.
In addition, they further reduced computation by explicitly estimating the historical IMU poses corresponding to only \textit{one} of the camera's imaging times, 
while using pose interpolation to represent the state at intermediate times corresponding to the other camera.
While in the proposed MIMC-VINS we leverage a similar interpolation scheme to reduce computation, we simultaneously perform time offset and spatial calibration between $n\geq 2$ cameras.

An integral part of any multi-sensor fusion system is the spatial (relative transformation), temporal (time offset), and intrinsic (e.g. focal length, camera center, rolling shutter readout, and distortion parameters) calibration parameters for each sensor, 
as errors in the values of these parameters can greatly degrade localization performance -- if not catastrophically.
Calibration can be broadly divided into two main categories.
Offline methods perform a computationally expensive solution process in exchange for providing highly accurate calibration estimates.
In particular, Furgale et al.~\cite{Furgale2013IROS} developed a multi-sensor calibration system that performed spatial, temporal, and intrinsic calibration of an arbitrary number of cameras along with an IMU.
However, performing offline calibration is a tedious process that limits deployment time and requires the calibration to be repeated if the sensor configuration changes.
In addition, treating the calibration parameters provided by these methods as ``known'' (zero uncertainty) may lead to unmodeled errors, thereby introducing estimation inconsistency~\cite{LiICRA2014}.
By contrast, online methods treat the calibration parameters as random variables with known priors and simultaneously estimate them along with the navigation states.
While many VINS algorithms perform online calibration of the spatial extrinsic transform between the camera and IMU, 
relatively few also estimate the time offset between them~\cite{Leutenegger2014IJRR,Sun2018RAL}.
Systems that \textit{do} perform online temporal calibration~\cite{Li2014IJRRa, Li2014IJRRb, Qin2018TRO}, however, are typically limited to a {\em single} IMU-camera pair.
As one of the most notably complete systems in this category, Li et al.~\cite{Li2014ICRA} performed online calibration of both the extrinsic parameters between a single IMU and camera as well as the \textit{intrinsics} of both sensors.
In our recent work~\cite{Yang2019RAL} we also have performed in-depth degenerate motion analysis to understand the effects of motions on sensor calibration.

\subsection{Multi-IMU Systems}

To date, almost all VINS algorithms utilize a single IMU, and thus these algorithms remain vulnerable to single IMU failure.
In reality, sensors certainly may experience failures preventing the estimator from acquiring new measurements from the faulty sensor.
If this sensor (such as IMUs in VINS) is required to fully constrain the estimation problem, its failure will result in the collapse of the entire system's ability to provide state estimates.
Such failures can occur in practice due to sensor disconnection (due to impact), high temperatures, or sensitivity to vibrations~\cite{avram2015imu}.
To compensate for this issue, redundant sensors (i.e. hardware redundancy) are typically used~\cite{Jeerage1990}.
As an added benefit, additional IMUs can improve localization accuracy by providing more information to the estimator.  
Therefore, adding more IMUs into VINS appears to be a straightforward solution for improving the system while additionally providing resilience against sensor failures,  
in particular, given the low cost of IMUs.
However, to the best of our knowledge, {\em few} VINS methods utilize {\em multiple} IMUs while performing real-time state estimation due to the challenges and complexity introduced.

While outside of VINS, fusing multiple IMUs has been studied~\cite{Bancroft2011DataFA, Jadid2019, Rasoulzadeh2016},
e.g. in the application to human motion tracking~\cite{Filippeschi2017}, 
these methods neither perform visual-inertial fusion nor online spatial/temporal calibration as in this work. 
Ma et al.~\cite{Ma2016} fused a tactical grade IMU, stereo camera, leg odometry, and GPS measurements in an EKF alongside a navigation-grade \textit{gyroscope} for estimating the motion of a quadruped robot, 
but without calibration or the ability to use acceleration measurements from a second IMU.

For offline calibration, Rehder et al. \cite{Rehder2016} used a continuous-time basis function representation~\cite{Furgale2013IROS} of the sensor trajectory 
to calibrate both the extrinsics and intrinsics of a multi-IMU system in a batch-based setting.
As this B-spline representation allows for the direct computation of expected local angular velocity and local linear acceleration,
the difference between the expected and measured inertial readings served as errors in the batch optimization formulation.
Kim et al. \cite{Kim2018} reformulated IMU preintegration~\cite{Eckenhoff2016WAFR, Forster2017TRO, Lupton2012TRO} by transforming the inertial readings from a first IMU frame into a second frame.
This allowed for spatial calibration with online initialization between an IMU and other sensors (including other IMUs), 
but did not include temporal calibration while also relying on computing angular accelerations from gyroscope measurements without optimal characterization of their uncertainty.

Recently, Zhang et al.~\cite{Zhang2019ALA} proposed a method for fusing multiple IMU's by setting up a ``virtual'' IMU and estimating its acceleration and angular velocity through least-squares using all the inertial measurements collected by every IMU. These synthetic readings, which have substantially smaller noises than those of each individual IMU, could then be used in VINS directly. While this was shown to offer competitive results along with large computational savings compared to the method proposed in our multi-IMU method~\cite{Eckenhoff2019ICRA_mi}, it requires \textit{perfectly} calibrated and synchronized sensors, which may be difficult to achieve in practice.
In addition, we note that because our system simply applies relative pose constraints to fuse the information, these updates can be used even if the calibration is \textit{time-varying}.

 \section{Preliminary: Single-Camera Single-IMU MSCKF-based VINS}  
\label{sec:vins}

In this section, we provide some background of the standard VINS with a single pair of calibrated camera and IMU,
by describing the IMU propagation and camera measurement models within the MSCKF framework~\cite{Mourikis2007ICRA},
which will serve as the basis of the proposed MIMC-VINS estimator.

Specifically, the state vector of the standard MSCKF-based VINS consists of the current IMU states 
and a sliding window of cloned IMU poses corresponding to the past $m$ images:\footnote{Throughout this paper 
the subscript $\ell |j$ refers to the estimate of a
quantity at time-step $\ell$, after all measurements up to time-step $j$ have been processed. $\hat
x$ is used to denote the estimate of a random variable $x$, while $\tilde x = x-\hat x$ is  the error in this estimate. 
$\mathbf I_n$ and $\mathbf 0_n$ are the $n \times n$ identity and zero matrices, respectively.
Finally, the left superscript denotes the frame of reference with respect to which the vector is expressed.}
\begin{align} 
\mathbf
x_k &= \begin{bmatrix} \mathbf x_{I,k}^\top & \mathbf x_{cl,k}^\top \end{bmatrix}^\top \label{eq:msckf-state} \\
\mathbf x_{I,k} &= \begin{bmatrix} ^I_G\bar{\mathbf q}_k^\top & \mathbf b_{g,k}^\top & ^G\mathbf v_{I,k}^\top & \mathbf b_{a,k}^\top & ^G\mathbf p_{I,k}^\top \end{bmatrix}^\top 
\label{eq:msckf-imu-state}\\
\mathbf{x}_{cl,k} &= 
\begin{bmatrix}
{}^I_G{\bar{\mathbf q}_k}^{\top} \!&\! {}^G\mathbf{p}_{I,k}^{\top} \!&\! \cdots \!&\! {}^I_G{\bar{\mathbf q}_{k-m+1}}^{\top} \!&\! {}^G\mathbf{p}_{I,{k-m+1}}^{\top}
\end{bmatrix}^{\top}
\label{eq:msckf-clone-state}
\end{align}
where $^I_G\bar{\mathbf q}$ is the unit quaternion that represents the rotation from the global frame of reference $\{G\}$ to the IMU frame $\{I\}$ 
(i.e. different parametrization of the rotation matrix $\mathbf R(^I_G\bar{\mathbf q}) =: {^I_G\mathbf R}$);
$^G\mathbf p_I$ and $^G\mathbf v_I$ are the IMU position and velocity in the global frame; $\mathbf b_g$ and $\mathbf b_a$ denote the gyroscope and accelerometer biases, respectively;
and $\{{}^I_G{\bar{\mathbf q}_{k-i}},  {}^G\mathbf{p}_{I,{k-i}} \}$ ($i=0,\cdots,m-1$) are the cloned IMU poses at time $t_{k-i}$.

\subsection{Propagation} \label{sec:imu-model}

The MSCKF propagates the state estimate 
based on the IMU continuous-time kinematics of the state~\eqref{eq:msckf-imu-state}~\cite{Trawny2005_Q_TR}:
\begin{align}
^I_G\dot{\bar{\mathbf q}}(t) &= \frac{1}{2} \bm\Omega\left(^I\bm\omega(t)\right) {^I_G\bar{\mathbf q}}(t),~ 
^G\dot{\mathbf p}_I(t) = {^G\mathbf v_I(t)} \notag\\
^G\dot{\mathbf v}_I(t) &= {^G\mathbf a}_I(t), ~
\dot {\mathbf b}_g(t) = \mathbf n_{wg}(t) ,~
\dot {\mathbf b}_a(t) = \mathbf n_{wa}(t) 
\label{eq:imu-motion}
\end{align}
where $^I\bm\omega = [ \omega_1 ~ \omega_2 ~ \omega_3 ]^\top$ is the rotational velocity of the IMU, 
expressed in $\{I\}$, $^G\mathbf a_I$ is the IMU acceleration in $\{G\}$, 
$\mathbf n_{wg}$ and $\mathbf n_{wa}$ are the white Gaussian noise processes that drive the IMU biases,
and $\bm\Omega(\bm\omega) = \begin{bmatrix}
-\lfloor \bm\omega \times \rfloor & \bm\omega \\
-\bm\omega^\top & 0 \end{bmatrix}$, where $\lfloor \bm\omega \times \rfloor$ is the skew-symmetric matrix.
A canonical 6-axis IMU provides gyroscope and accelerometer measurements, $\bm\omega_m$ and $\mathbf a_m$, 
both of which are expressed in the IMU local frame $\{I\}$ and at time-step $t_k$ are given by:
\begin{align}
\bm\omega_m(t_k) &= {^I\bm\omega}(t_k) + \mathbf b_g(t_k) + \mathbf n_g(t_k) \label{eq:imu-wm}\\
\mathbf a_m(t_k) &= 
\scalemath{.9}{
^I_G\mathbf R(t_k)
\left( {^G\mathbf a}_I(t_k) + {^G\mathbf g}  \right) + \mathbf b_a(t_k) + \mathbf n_a(t_k)  }
\label{eq:imu-am}
\end{align}
where ${^G\mathbf g}$ is the gravitational acceleration expressed in $\{G\}$, 
and $\mathbf n_g$ and $\mathbf n_a$ are zero-mean white Gaussian noise.
Using the inertial measurements collected between the time interval $[t_k,t_{k+1}]$ [see \eqref{eq:imu-wm} and \eqref{eq:imu-am}], denoted by $\mathbf u_m(t_k:t_{k+1})$,
and based on the above kinematic model~\eqref{eq:imu-motion}, we can propagate (via integration) the IMU state in discrete time~\cite{Chatfield1997}:
\begin{align}
\hat{\mathbf x}_I(t_{k+1}) &= \mathbf f (\hat{\mathbf x}_I(t_{k}), \mathbf u_m(t_k:t_{k+1}), \mathbf 0)
\label{eq:imu-state-est-prop}
\end{align}
where last entry, $\mathbf 0$, corresponds to the zero-mean noise vector.

To propagate the covariance matrix, 
we first define the error state as follows [see~\eqref{eq:msckf-state}]:
\begin{align} \label{eq:imu-err-state}
\scalemath{.9}{
\widetilde{\mathbf x}(t) = 
\begin{bmatrix}
{}^{I(t)}_G\widetilde{\bm\theta}^\top  \!&\! \mathbf{\widetilde b}_g^\top(t) \!&\!  ^G\mathbf{\widetilde v}_I^\top(t) \!&\!  \mathbf{\widetilde b}_a^\top(t) 
\!&\! ^G\mathbf{\widetilde p}_I^\top(t) \!&\! \mathbf{\widetilde x}_{cl}^\top(t)
\end{bmatrix}^\top
}
\end{align}
where we have employed the multiplicative error model for a quaternion~\cite{Trawny2005_Q_TR}.
That is, the error between the quaternion $\bar{\mathbf q}$ and its estimate $\hat{\bar{\mathbf q}}$ is the $3\times 1$ angle-error vector,
$\widetilde{\bm\theta}$, implicitly defined by the error quaternion: 
$\delta\bar{\mathbf q} = \bar{\mathbf q} \otimes \hat{\bar{\mathbf q}} \simeq [
\frac{1}{2} {\widetilde{\bm\theta}}^\top ~ 1 ]^\top$,
where $\delta\bar{\mathbf q}$ describes the small rotation that causes the true and estimated attitude to coincide.
Then, linearizing~\eqref{eq:imu-motion} at the current state estimate yields 
the following continuous-time error-state propagation:
\begin{align}
\dot{\widetilde{\mathbf x}}(t) = 
\mathbf F_c(t) \widetilde{\mathbf x}(t) + \mathbf G_c(t) \mathbf n(t)
\end{align}
where 
$\mathbf n = [ \mathbf n_g^\top ~ \mathbf n_{wg}^\top ~ \mathbf n_a^\top ~ \mathbf n_{wa}^\top ]^\top$
is the system noise, 
$\mathbf F_c$ is the continuous-time error-state transition matrix, 
and $\mathbf G_c$ is the input noise matrix~\cite{Trawny2005_Q_TR}.
The system noise is modeled as zero-mean white Gaussian process with autocorrelation 
$\mathbb E [ \mathbf n(t) \mathbf n(\tau)^\top ] = \mathbf Q_c \delta(t-\tau)$,
which depends on the IMU noise characteristics.
Based on this continuous-time propagation model using IMU measurements,
the discrete-time state-transition matrix, $\bm\Phi_k := \bm\Phi(t_{k+1},t_{k})$, 
is required in order to propagate the error covariance from time $t_{k}$ to $t_{k+1}$.
Typically it is found by solving the following matrix differential equation:
\begin{align} \label{eq:mat-diff}
\dot{\bm\Phi} (t,t_k) = \mathbf F_c(t) \bm\Phi (t,t_k)
\end{align}
with the initial condition $\bm\Phi(t_{k},t_k) = \mathbf I_{15+6m}$.
This can be solved either numerically~\cite{Mourikis2009TRO} 
or analytically~\cite{Hesch2013TRO,Li2012ICRA,Li2013IJRR}.
Once it is computed, 
the MSCKF propagates the covariance as in the standard EKF~\cite{Maybeck1979}:
\begin{align} \label{eq:cov-prop}
\mathbf P_{k+1|k} = \bm\Phi \left(t_{k+1}, t_k \right) \mathbf P_{k|k} \bm\Phi \left(t_{k+1}, t_k \right)^\top + \mathbf Q_{d,k}
\end{align}
where $\mathbf Q_{d,k}$ is the discrete-time system noise covariance:
\begin{align}
\scalemath{.95}{\mathbf Q_{d,k} = \int_{t_{k}}^{t_{k+1}} \bm\Phi(t_{k+1},\tau) \mathbf G_c(\tau) \mathbf Q_c \mathbf G_c^\top(\tau)  \bm\Phi^\top(t_{k+1},\tau) d\tau }
\end{align}
Note that after propagation the MSCKF performs stochastic cloning~\cite{Roumeliotis2002ICRAa} to probabilistically augment the state vector, $\mathbf{x}_{cl,k}$, and covariance matrix with the current pose estimate.

\subsection{Update} \label{sec:meas-model}

We detect and track a point feature over images and use its measurements to update the state estimate and covariance.
Specifically, assuming a calibrated perspective camera, the normalized measurement of this feature at time $t_j$ is 
the perspective projection of its 3D position in the camera frame, $^{C_j}\mathbf p_{f} = [ x_j ~ y_j ~ z_j ]^\top$,
onto the image plane (which is normalized with known camera intrinsics), for $j=k+1,\cdots,k-m$:
\begin{align} \label{eq:meas-model}
\mathbf z_{n,j} &= \bm\Pi \left({}^{C_j}\mathbf{p}_f\right) + \mathbf n_{z,j}  = \frac{1}{z_j} \begin{bmatrix} x_j \\ y_j \end{bmatrix} + \mathbf n_{z,j} \\
{^{C_j}\mathbf p_{f}} &= 
{^C_I\mathbf R} {^{I_j}_G\mathbf R} \left({^G\mathbf p_{f}} - {^G\mathbf p_{I,j}} \right) + {^C\mathbf p_I}
\label{eq:meas-eq}
\end{align}
where ${^G\mathbf p_{f}}$ is the 3D global position of the observed feature, and $\mathbf n_{z,j}$ is the zero-mean white Gaussian measurement noise;
$\{{^C_I\mathbf R},  {^C\mathbf p_I} \}$ is the rotation and translation between the camera and IMU, which can be obtained, for example, 
by performing camera-IMU extrinsic calibration {\em offline}~\cite{Mirzaei2008TRO,Furgale2013IROS}.
As the feature is not kept in the MSCKF state vector [see \eqref{eq:msckf-state}], 
in order to perform EKF update with the above measurement~\eqref{eq:meas-model},
we first perform bundle adjustment (BA) using all its measurements available in the current window (while fixing the camera pose estimates) to obtain the linearization point ${^G\hat{\mathbf p}_f}$~\cite{Mourikis2007ICRA}.
Linearization of~\eqref{eq:meas-model} around the current state estimate and this feature linearization point
yields the following measurement residual [see~\eqref{eq:imu-err-state}]:
\begin{align} \label{eq:residual}
\widetilde{\mathbf z}_{n,j} &= \mathbf H_{x,j} \widetilde{\mathbf x}_{{k+1|k}}  + \mathbf H_{f,j} {^G\widetilde{\mathbf p}_{f}} + \mathbf n_{z,j}
\end{align}
By stacking all these measurement residuals for $j=k+1,\cdots,k-m$, 
we perform the nullspace operation (linear marginalization of the feature~\cite{Yang2017IROS}) 
to infer the new measurement residual whose noise is independent of the state:
\begin{align} 
&\scalemath{.9}{\underbrace{\begin{bmatrix} \widetilde{\mathbf z}_{n,k+1}\\ \vdots \\ \widetilde{\mathbf z}_{n,k-m} \end{bmatrix} }_{\widetilde{\mathbf z}}
= \underbrace{\begin{bmatrix} {\mathbf H}_{x,k+1}\\ \vdots \\ {\mathbf H}_{x,k-m} \end{bmatrix}}_{{\mathbf H}_{x}}  \widetilde{\mathbf x}_{{k+1|k}}  + 
\underbrace{ \begin{bmatrix} {\mathbf H}_{f,k+1}\\ \vdots \\ {\mathbf H}_{f,k-m} \end{bmatrix} }_{\mathbf H_f}  {^G\widetilde{\mathbf p}_{f}} + 
\underbrace{ \begin{bmatrix} {\mathbf n}_{z,k+1}\\ \vdots \\ {\mathbf n}_{z,k-m} \end{bmatrix} }_{\mathbf n_z} } \notag\\
&\overset{\mathbf N^\top \mathbf H_f = \mathbf 0}{\Longrightarrow}~~
\mathbf N^\top \widetilde{\mathbf z}  =  \mathbf N^\top  \mathbf H_x  \widetilde{\mathbf x}_{{k+1|k}}  + \mathbf N^\top  \mathbf n_{z}
\label{eq:msckf-residual}
\end{align}
which can now be used for the standard EKF update~\cite{Mourikis2007ICRA}.

 \section{Barebones  Multi-IMU Multi-Camera (MIMC)-VINS} 
\label{sec:mimc-vins}

Building upon our recent multi-IMU~\cite{Eckenhoff2019ICRAb} and multi-camera VINS~\cite{Eckenhoff2019ICRAa},
we integrate both functionalities into a tightly-coupled MSCKF-based multi-IMU multi-camera (MIMC)-VINS.
In this section, 
we present in detail the {\em barebones} MIMC-VINS by assuming all sensors are calibrated,
and will later extend this system to uncalibrated, and possibly faulty sensors.

Consider a fully calibrated sensor platform (i.e. the extrinsic parameters between them are known) consisting of $N+1$ IMUs and $M+1$ cameras.
We will develop an efficient MSCKF-based VINS estimator to fuse all information provided by these sensors,
which has the following state vector [see \eqref{eq:msckf-state}]:
\begin{align} \label{eq:mimc-state-bear}
\mathbf x_k = \begin{bmatrix}
\mathbf x_{I,k}^\top & \mathbf x_{cl,k}^\top 
\end{bmatrix}^{\top}   
= \begin{bmatrix}
\mathbf{x}_{I_0,k}^{\top} &
\cdots &
\mathbf{x}_{I_N,k}^{\top} & \mathbf x_{cl,k}^\top 
\end{bmatrix}^{\top}  
\end{align}
where $\mathbf{x}_{I_i,k}$ is the navigation state of the $i$-th IMU [see~\eqref{eq:msckf-imu-state}].
To allow for utilizing visual feature measurements, 
we select an {\em arbitrary} IMU to serve as the ``base'', denoted by $\{I_b\}$, 
which can be changed over the trajectory if needed (see Section~\ref{sec:sensor-fail}). 
As in the standard MSCKF, we keep a sliding window of stochastically cloned poses of only this {\em base} IMU.
Specifically, at time $t_k$
we also maintain a sliding window of the base IMU clones at $m$ past imaging times $t_j$ ($j=k,\cdots,k-m+1$) of the base camera (which is also arbitrarily chosen)
[see \eqref{eq:msckf-clone-state}]:
\begin{align}  \label{eq:clone-state-imu}
    &\mathbf{x}_{cl,k}= \\
    &\begin{bmatrix}
       {}^{I_b(t_{k})}_G\bar{\mathbf{q}}^{\top} \!&\! {}^G\mathbf{p}_{I_b(t_{k})}^{\top} \!&\! \cdots \!&\! {}^{I_b(t_{k-m+1})}_G\bar{\mathbf{q}}^{\top} \!&\! {}^G\mathbf{p}_{I_b(t_{k-m+1})}^{\top}
    \end{bmatrix}^{\top} \notag
\end{align}
Note that in the above we have used a slightly different notation from \eqref{eq:msckf-clone-state} to highlight the exact times of cloned poses.

\subsection{Propagation}

As all $N+1$ IMUs' states are included in the state vector~\eqref{eq:mimc-state-bear},
we propagate each IMU's state estimate and the joint covariance using each IMU's measurements as in the standard MSCKF.
Specifically, 
at the current time step $k$ (corresponding to the current imaging time of the base camera $t_k$),
we propagate the current IMU state estimate forward to the next time step $k+1$ (corresponding to the new image time $t_{k+1}$), 
by using all the IMU measurements available in the time window $[t_k, t_{k+1}]$ as in~\eqref{eq:imu-state-est-prop} for each IMU.
Similarly, the error covariance can be propagated as follows [see \eqref{eq:cov-prop}]:
\begin{align}
\mathbf{P}_{k+1|k} &= \bm \Phi_k  \mathbf{P}_{k|k} \bm \Phi^{\top}_k+ \mathbf{Q}_{d,k} \label{eq::big_prop}\\
\bm \Phi_k &= 
\scalemath{.95}{
\mathbf{Diag}\left(\bm \Phi_0(t_{k+1}, t_k), \hdots, \bm \Phi_N(t_{k+1}, t_k), \mathbf{I}_{6m} \right)
} \label{eq::big_phi}\\
\mathbf{Q}_{d,k} &= \mathbf{Diag}\left( \mathbf{Q}_{0d,k}, \hdots, \mathbf{Q}_{Nd,k}, \mathbf{0}_{6m} \right) 
\end{align}
where $\bm \Phi_i(t_{k+1}, t_k)$ is the linearized state-transition matrix for the error state of the $i$-th IMU across the time interval $[t_k, ~t_{k+1} ]$ 
and $\mathbf{Q}_{id}$ is the corresponding noise covariance, while $\mathbf{Diag}(\cdots )$ places the argument matrix entries on the block diagonals of an otherwise zero matrix.  
Each of these matrices are computed \textit{per IMU} using its measurements as in the standard single-IMU/camera case.
In the above expressions, the identity matrix $\mathbf I_{6m}$ in $\bm\Phi_k$ 
and the right-bottom zero matrix $\mathbf 0_{6m}$ in $\mathbf Q_{d,k}$ correspond to the states with zero dynamics.
It is important to note that we are not estimating the state of each IMU independently, 
and instead allow for tracking of all cross correlations between IMUs,
which are important for optimal fusion of multi-sensor measurements. We also note that when computing~\eqref{eq::big_prop} we take advantage of the sparse, block-diagonal structure of~\eqref{eq::big_phi} to perform the computation efficiently.

\subsection{Update}

In contrast to the single-IMU/camera scenario, 
in the proposed MIMC-VINS we perform efficient EKF update using both image feature tracks \textit{and} multi-IMU relative pose constraints.

\subsubsection{Multi-IMU Constraints}

As all IMUs are rigidly connected, at any time $t$ we have the following relative transformation between the base and non-base IMUs:
\begin{align}
    {}^{I_b}_{I_i}\bar{\mathbf{q}} &= {}^{I_b(t)}_{G}{\bar{\mathbf{q}}} \otimes {}^{I_i(t)}_{G}{\bar{\mathbf{q}}}^{-1} \\
   {}^{I_b}\mathbf{p}_{I_i}  &= {}^{I_b(t)}_G\mathbf{R}\left({}^G\mathbf{p}_{I_i}(t) -{}^G\mathbf{p}_{I_b}(t)\right) 
\end{align}
where ${}^{I_b}_{I_i}\bar{\mathbf{q}}$ and ${}^{I_b}\mathbf{p}_{I_i}$ are the {\em fixed} relative pose (extrinsic calibration) between the base IMU $b$ (say $b=0$) 
and the $i$-th IMU (say $i=1,\cdots,N$).
The residual associated with this constraint for each IMU can be written as:
\begin{align} 
\begin{cases}
    2\textbf{vec}\left({}^{I_b(t)}_{G}{\bar{\mathbf{q}}} \otimes {}^{I_i(t)}_{G}{\bar{\mathbf{q}}}^{-1}\otimes {}^{I_b}_{I_i}\bar{\mathbf{q}}^{-1}\right)&= \mathbf{0} 
\\
      {}^G\mathbf{p}_{I_b}(t) + {}^G_{I_b(t)}\mathbf{R}{}^{I_b}\mathbf{p}_{I_i}- {}^G\mathbf{p}_{I_i}(t)&= \mathbf{0} \label{eq:res_const_pos} 
    \end{cases}~
    \Rightarrow \mathbf{r}_{I_i}\left(\mathbf{x}\right) &= \mathbf{0}
\end{align}
where $\textbf{vec}(\bar{\mathbf{q}})= \mathbf{q}$ returns the $3\times 1$ vector portion of the argument quaternion $\bar{\mathbf q}$. 
We stack this constraint for each auxiliary IMU to form a system of residuals, $\mathbf{r}_I(\mathbf{x})=\mathbf 0$.
Linearization of these residuals at the current  estimate yields:
\begin{align} \label{eq:mi-residual}
\mathbf{r}_I\left(\hat{\mathbf{x}}\right)+\frac{\partial \mathbf{r}_I}{\partial \tilde{\mathbf{x}}}\tilde{\mathbf{x}} &\approx \mathbf{0} 
~\Rightarrow~
\mathbf{0} -\mathbf{r}_I\left(\hat{\mathbf{x}}\right) \approx \frac{\partial \mathbf{r}_I}{\partial \tilde{\mathbf{x}}}\tilde{\mathbf{x}}
\end{align}

Note that this is a \textit{hard} constraint that acts as a measurement with \textit{zero} noise. Alternatively, we may loosen this constraint by adding a small, synthetic noise to this measurement. Such a treatment may be useful to handle unmodeled errors such as flexible calibration.
In practice, these measurements may quickly degrade performance due to inaccuracies in the calibrated transforms between sensors,
which motivates us to perform online calibration of these parameters in the next section.

It should be noted that such relative-pose constraints have been used in previous multi-IMU systems~\cite{Bancroft2009}, along with a constraint on the relationship between the IMUs' velocities. 
However, transferring velocities from one frame to another requires angular velocity measurements, which have already been used in the propagation step,
and would therefore introduce unmodeled correlations between the propagation and update noises.
As such, we choose to forgo this constraint to ensure consistency.
In addition, 
we utilize these relative-pose constraints to perform both spatial and temporal calibration of the sensors as shown in the next section.
While in this work we assume the spatial calibration parameters remain static, i.e. the sensors are rigidly mounted, they can also be modeled as random walks when dealing with a more flexible mount and can be directly introduced in the above framework with ease.

\subsubsection{Multi-Camera Measurements}

\begin{figure}
\centering
\includegraphics[width=0.6\columnwidth]{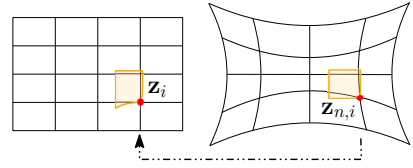}
\caption{Illustration of undistorting from normalized to a raw image pixel.}
\label{fig:undistort}
\end{figure}

Consider a 3D point feature, ${}^G\mathbf{p}_f$, that is captured by the $i$-th camera at imaging time $t$.
This measurement function is given by [see \eqref{eq:meas-model}]:
\begin{align} \label{eq:feature_in_cam0}
    \mathbf{z}_{i}(t)&= \mathbf{w}_i\left(\bm \Pi \left({}^{C_i\left(t\right)}\mathbf{p}_f\right),~ \bm \zeta_i \right)+ \mathbf{n}_{z_i}(t) \\
   {}^{C_i\left(t\right)}\mathbf{p}_f &= {}^{C_i}_I\mathbf{R}{}^{I(t)}_G\mathbf{R}\left({}^G\mathbf{p}_f-{}^G\mathbf{p}_{I(t)}\right)+{}^{C_i}\mathbf{p}_I 
    \label{eq:feature_in_cam}
\end{align}
where $\mathbf{w}_i(\cdot)$ is the function mapping the normalized image coordinates $\mathbf {z}_{n,i}$~\eqref{eq:meas-model}
onto the image plane based on the camera intrinsics $\bm\zeta_i$ and the camera model used (e.g., radial-tangential or fisheye~\cite{OPENCV_library}), 
and ${}^{C_i\left(t\right)}\mathbf{p}_f$ is the position of the feature expressed in the $i$-th camera frame at time $t$ [see also \eqref{eq:meas-eq}].
Fig.~\ref{fig:undistort} visualizes this image distortion operation.
Note that rather than using the undistorted, normalized pixel coordinates as measurements as in~\eqref{eq:meas-model}, 
we here model the \textit{raw} image coordinates,
which depend on the intrinsics of each camera $\bm\zeta_i$ that may include the focal lengths, principal point, and distortion parameters~\cite{Hartley2004}.
Importantly, this measurement model also allows us to calibrate online all cameras' intrinsic parameters as presented later in Section~\ref{sec:mimc-online-calib}.

\begin{figure}
\centering
\includegraphics[width=.7\columnwidth]{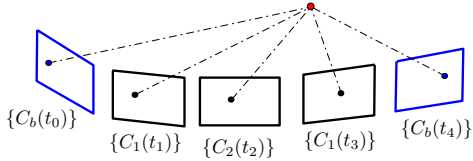}
\caption{Illustration of how asynchronous multi-camera measurements are collected. 
We have cloned at the base camera imaging times: $\{C_b(t_0)\}$, $\{C_b(t_4)\}$ (blue). 
A series of measurements between these times from non-base cameras $C_1$ and $C_2$ are received.
We can interpolate to these pose times using the base camera cloned poses.
}
\label{fig:async_meas_diagram}
\end{figure}

It is important to note that 
the cloned state $\mathbf{x}_{cl}$~\eqref{eq:clone-state-imu}
only contains the base IMU poses at the imaging times of the \textit{base} camera, 
while the camera measurements~\eqref{eq:feature_in_cam0} require that for all non-base cameras we can express the pose at \textit{their} image times, which may not align with those of the base camera (see Figure \ref{fig:async_meas_diagram}). 
Clearly, 
without solving this issue,
the inclusion of all other measurements that are not collected synchronously with the base camera cannot be written as functions of the state and used in the MSCKF update.
Naively, one could simply perform stochastic cloning at \textit{every} imaging time, however this would lead to a greatly increased computational burden. 
For instance, utilizing $m$ distinct asynchronous cameras each operating at the same frequency would require multiplying the state size by $m$ due to the large number of required clones.
Therefore, we instead employ $SO(3)\times \mathbb{R}^3$ on-manifold linear interpolation between these poses to allow for the incorporation of measurements at arbitrary times.

Assuming the imaging time of the $i$-th camera, $t$,
we let $t_1$ and $t_2$ denote the bounding base camera/IMU clones times which
$t$
falls between.
We linearly interpolate the cloned poses at $t_1$ and $t_2$ to find the pose at the measurement time:
\begin{align} \label{eq:linear-intpl-1}
    {}^{I(t)}_G\mathbf{R}&= \textrm{Exp}\left(\lambda\textrm{Log}\left({}^{I(t_2)}_{G}\mathbf{R}{}^{G}_{I(t_1)}\mathbf{R}\right)\right){}^{I(t_1)}_{G}\mathbf{R} \\ 
    {}^G\mathbf{p}_{I(t)}&= (1-\lambda){}^G\mathbf{p}_{I(t_1)}+ \lambda {}^G\mathbf{p}_{I(t_2)}
    \label{eq:linear-intpl-2}
\end{align}
where $\lambda= (t-t_1)/(t_2-t_1)$, 
and $\textrm{Log}(\cdot)$ is the inverse operation of $\textrm{Exp}(\cdot)$ which maps a rotation matrix to a vector in $\mathbb{R}^3$ \cite{chirikjian2011stochastic}.
These equations essentially interpolate both the orientation and position under the approximation of {\em constant} linear and angular velocity over the interval.
This may serve as a good approximation in cases where the base camera arrives at high rates (e.g. around 20 Hz) as compared to the physical camera motion.
 In addition, due to the fact that marginalization is only ever performed on the oldest clone, any required camera pose is typically within 25 milliseconds of a clone.
Note that a similar interpolation scheme was also used in~\cite{Paul2018CVPR} to reduce the complexity of processing a synchronized stereo pair.
Nevertheless, such linear interpolation may not be accurate for fast motions, 
which will be addressed by employing high-order polynomial interpolation introduced in Section \ref{sec:HOT_interp}.

As evident, the above linear interpolation [see \eqref{eq:linear-intpl-1} and \eqref{eq:linear-intpl-2}]
causes the visual measurement [see~\eqref{eq:feature_in_cam0} and \eqref{eq:feature_in_cam}] to be dependent on the base IMU clones that are in the state vector [see \eqref{eq:clone-state-imu}], resulting in nontrivial computation of measurement Jacobians. 
Specifically, 
let $\bm \eta(t_1)$,  $\bm \eta(t_2)$ and $\bm \eta(t)$ be the IMU cloned poses, $\bm \eta = \{ {}^{I}_G\mathbf{R}$, ${}^G\mathbf{p}_{I} \}$, at the neighboring times and the interpolated value, respectively.
By substituting~\eqref{eq:linear-intpl-1} and \eqref{eq:linear-intpl-2} into the measurement function~\eqref{eq:meas-eq} and thus \eqref{eq:feature_in_cam},  
we have the following Jacobians with respect to the bounding IMU clones using the chain rule of differentiation:
\begin{align}
\frac{\partial \mathbf {\tilde z}_i(t)}{\partial \tilde{\bm \eta}(t_1) }   &= 
\frac{\partial \mathbf {\tilde z}_i(t)}{\partial \mathbf {\tilde z}_{ni}(t) } 
\frac{\partial \mathbf {\tilde z}_{ni}(t) } {\partial {}^{C_i(t)}\tilde{\mathbf{p}}_f }
\frac{\partial{}^{C_i(t)}\tilde{\mathbf{p}}_f }{\partial \tilde{\bm \eta}(t)}\frac{\partial \tilde{\bm \eta}(t)}{\partial \tilde{\bm \eta}(t_1)} \\
\frac{\partial \mathbf {\tilde z}_i(t)}{\partial \tilde{\bm \eta}(t_2) }   &= 
\frac{\partial \mathbf {\tilde z}_i(t)}{\partial \mathbf {\tilde z}_{ni}(t) } 
\frac{\partial \mathbf {\tilde z}_{ni}(t) } {\partial {}^{C_i(t)}\tilde{\mathbf{p}}_f }
\frac{\partial{}^{C_i(t)}\tilde{\mathbf{p}}_f }{\partial \tilde{\bm \eta}(t)}\frac{\partial \tilde{\bm \eta}(t)}{\partial \tilde{\bm \eta}(t_2)}
\end{align} 
In computing these Jacobians, while the derivatives (the first three terms) of image raw pixels with respect to the interpolated pose are straightforward,
the last term $\frac{\partial \tilde{\bm \eta}(t)}{\partial \tilde{\bm \eta}(t_i)}$ ($i=1,2$) requires derivatives of the interpolation function with respect to the bounding clones.
For brevity, we include the  detailed derivations for general interpolation in Appendix~\ref{appendix::HOI}.

\subsubsection{Parallelizing Visual Tracking}
\label{sec:parallel-tracking}

\begin{figure}
\centering
\includegraphics[width=0.6\columnwidth]{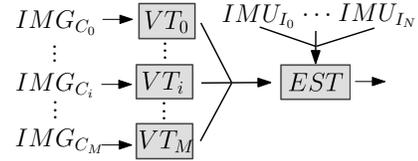}
\caption{Illustrate how the proposed system horizontally scales as more images are added. Simply scaling of the visual tracker (VT) allows for the parallelization of feature tracking, which feeds these tracks to the centralized estimator (EST) for processing.}
\label{fig:parallel}
\end{figure}

A key advantage of the proposed MIMC-VINS estimator is the ability to parallelize such visual tracking front-end.
Since all cameras can be processed independently, we argue that one can perform ``camera-edge'' visual tracking allowing for the horizontal scaling of the visual front-ends.
As seen in Figure \ref{fig:parallel}, the visual front-ends can be treated as independent and images from all cameras can be processed in parallel.
For example, in practice one could let each camera have a local micro-computer or hardware embedded processor (``camera-edge'' processing) that performs feature tracking that upon completion can be sent to the centralized estimator for asynchronous fusion.
In this work, we perform simple multi-threaded optimization such that each camera has its own thread to perform feature extract, tracking, and outlier rejection.

 \section{Advanced MIMC-VINS with Online Calibration}
\label{sec:mimc-ext}

To advance the barebones MIMC-VINS presented in the preceding section, 
the proposed VINS estimator is able to:
(1) perform online calibration of both intrinsic and spatiotemporal extrinsic parameters to enable plug-and-play functionality, thus lowering the technology barriers for end users;
(2) enforce VINS observability constraints in computing filter Jacobians to improve estimation consistency;
and (3) offer smooth, uninterrupted, resilient estimates when faced with sensor failures.

\subsection{Incorporating Online Sensor Calibration}
\label{sec:mimc-online-calib}

Treating the spatial extrinsic calibration parameters (i.e. the 6DOF rigid-body transformation) as known may lead to large errors which affect the estimation performance, in particular, when working with low-cost sensor systems (e.g. \cite{Li2014ICRA}). 
To combat this, we model these quantities as random variables and add them into our filtering framework.
Moreover, {\em asynchronous} independent IMUs with non-negligible time offsets relative to the base IMU clock (which can be arbitrarily chosen, say $b=0$), along with uncertain rigid IMU to IMU transformations can greatly impact the multi-IMU constraints~\eqref{eq:res_const_pos} which are crucial to ensuring high quality state estimation and sensor resiliency.
Thus, we are motivated to perform online estimation of these parameters.
This involves storing these parameters in our static calibration state and computing the Jacobians with respect to these quantities whenever they appear in our measurement functions.

We model each IMU's time offset from the clock of the base IMU, which may be due to hardware or data transfer latency. In particular, consider a time ${}^{I_i}t$ as expressed in the $i$-th IMU's clock, which is related to the same instant represented in the base IMU clock, ${}^{I_b}t$, by a time offset ${}^{I_b}t_{I_i}$:
\begin{align} \label{eq:timeoffset-imus}
    {}^{I_i}t= {}^{I_b}t+{}^{I_b}t_{I_i}
\end{align}
where ${}^{I_b}t_{I_i}$ is estimated online.
Similarly, the time reported by a (base) camera will {\em differ} from the same time expressed in the base IMU's clock 
by a time offset ${}^{C_b}t_{I_b}$:
\begin{align} \label{eq:imu-cam-timeoffset}
    {}^{I_b}t = {}^{C_b}t+ {}^{C_b}t_{I_b}
\end{align}
Lastly, we also need to estimate the time offset ${}^{C_i}t_{C_b}$ between the $i$-th camera's clock and the base camera  (which can also be arbitrarily chosen)
in order to properly fuse the multi-camera measurements:
\begin{align} \label{eq:cam-cam-timeoffset}
    {}^{C_b}t = {}^{C_i}t+ {}^{C_i}t_{C_b}
\end{align}

From this, we augment our MIMC-VINS state vector~\eqref{eq:mimc-state-bear} to include our spatiotemporal extrinsics between the $N+1$ IMUs and $M+1$ cameras as well as every cameras' intrinsic parameters [see \eqref{eq:feature_in_cam0}]:
\begin{align}\label{eq:mimc-w-cal-state}
\mathbf x_k &=
 \begin{bmatrix}
\mathbf{x}_{I_b,k}^{\top} &
\mathbf{x}_{I_1,k}^{\top} &
\cdots &
\mathbf{x}_{I_N,k}^{\top} & \mathbf x_{cl,k}^\top 
& \mathbf x_{cal,k}^\top \end{bmatrix}^{\top} \\
\mathbf x_{cal,k} &= \begin{bmatrix} \mathbf{x}_{e1}^\top &\cdots & \mathbf{x}_{eN}^\top & \mathbf{x}_{cb}^\top & \mathbf{x}_{c1}^\top & \cdots & \mathbf{x}_{cM}^\top \end{bmatrix}^{\top}\\
\mathbf{x}_{ei} &= \begin{bmatrix} {}^{I_b}_{I_i}\bar{\mathbf{q}}^{\top} & {}^{I_b}\mathbf{p}_{I_i}^{\top} & {}^{I_b}t_{I_i} \end{bmatrix}^{\top}\\
\mathbf{x}_{cb} &= \begin{bmatrix} {}^{C_b}_{I_b}\bar{\mathbf{q}}^{\top} & {}^{C_b}\mathbf{p}_{I_b}^{\top} & {}^{C_b}t_{I_b} & \bm \zeta_b^{\top} \end{bmatrix}^{\top} \\
\mathbf{x}_{ci} &= \begin{bmatrix} {}^{Ci}_{I_b}\bar{\mathbf{q}}^{\top} & {}^{C_i}\mathbf{p}_{I_b}^{\top} & {}^{C_i}t_{C_b} & \bm \zeta_i^{\top} \end{bmatrix}^{\top} \\
\bm \zeta_{b/i} &= \begin{bmatrix} f_{xi} & f_{yi} & p_{xi} & p_{yi} & \mathbf{d}_i^{\top} \end{bmatrix}^{\top} 
\end{align}
where $\mathbf{x}_{ei}$ is the spatial and temporal calibration parameters  between $i$-th and the base IMUs;
that is, the relative pose between each auxiliary IMU and the base, ${}^{I_b}\mathbf{x}_{I_i} = [ {}^{I_b}_{I_i}\bar{\mathbf{q}}^{\top} ~ {}^{I_b}\mathbf{p}_{I_i}^{\top} ]^{\top}$,
and the the time offsets between them, ${}^{I_i}t_{I_b}$;
$\mathbf{x}_{cb}$ is the base camera's calibration parameters including 
the spatiotemproal extrinsics between the base camera and the base IMU and the camera intrinsic parameters $\bm \zeta_b$.
and similarly, $\mathbf{x}_{ci}$ is the calibration parameters for the $i$-th camera where we estimate the time offset between the $i$-th camera and the base camera.
For the camera intrinsics $\bm\zeta_{b/i}$, 
$f_{xi},~ f_{yi}$ represent the focal lengths, $p_{xi},~p_{yi}$ denote the location of the principal point, and $\mathbf{d}_i$ refers to the vector of distortion parameters whose length/definition depends on the camera model being used (see \cite{Hartley2004}).

In what follows we present in detail the key modifications required in each of the main steps of the proposed MIMC-VINS when jointly estimating the above calibration parameters $\mathbf x_{cal}$ within the MSCKF framework.

\subsubsection{Propagation}

We propagate the {\em base IMU} ($b=0$)  state  in analogy to~\cite{LiICRA2014}.
With the estimate of the time offset ${}^{C_b}\hat{t}_{I_b}$,
whenever we receive the $(k+1)$-th image with reported time ${}^{C_b}t_{k+1}$ (corresponding to the discrete time step $k+1$), 
we perform propagation of the base IMU up to the \textit{estimated} time of the image as expressed in the base IMU clock [see~\eqref{eq:imu-cam-timeoffset}]: 
${}^{I_b}\hat{t}_{k+1}= {}^{C_b}t_{k+1}+ {}^{C_b}\hat{t}_{I_b}$.
Specifically, we propagate the base IMU from its current time ${}^{I_b}\hat{t}_k$ 
up to this new time by processing all the base IMU measurements  collected over the time interval $[{}^{I_b}\hat{t}_k, {}^{I_b}\hat{t}_{k+1}]$ [see \eqref{eq:imu-state-est-prop}]:
\begin{align} \label{eq:mimc-prop000}
\mathbf{\hat x}_{I_0}({}^{I_b}\hat{t}_{k+1}) =\mathbf{f}\Big(\mathbf{\hat x}_{I_0}({}^{I_b}\hat{t}_k), \mathbf u_{m0} ({}^{I_b}\hat{t}_k: {}^{I_b}\hat{t}_{k+1} ), \mathbf 0 \Big)
\end{align}

Similarly, we propagate each {\em auxiliary} IMU state to the same base IMU time,
in order to correctly enforce the asynchronous multi-IMU relative pose constraints~\eqref{eq:res_const_pos}.
Specifically,
we propagate the $i$-th IMU's ($i=1,\cdots,N$) state up to the \textit{estimate} of the current base IMU time as expressed in the $i$-th IMU clock,
${}^{I_i}\hat{t} = {}^{I_b}\hat t+ {}^{I_b}\hat{t}_{I_i}$: 
\begin{align}\label{eq:mimc-propiii}
\mathbf{\hat x}_{I_i}({}^{I_i}\hat{t}_{k+1}) =\mathbf{f}\Big(\mathbf{\hat x}_{I_i}({}^{I_i}\hat{t}_k), \mathbf u_{mi} ({}^{I_i}\hat{t}_k: {}^{I_i}\hat{t}_{k+1} ), \mathbf 0 \Big)
\end{align}

As a result, the IMU state after propagation at time step $k+1$ can be written as (noting again $b=0$): 
\begin{align} \label{eq:mimc-imu-state-prop}
    {\mathbf x}_I({}^{I_b}\hat t_{k+1}) = 
       \begin{bmatrix}
    \mathbf{ x}_{I_0}({}^{I_0}\hat{t}_{k+1})^\top &
    \cdots &
    \mathbf{ x}_{I_N}({}^{I_N}\hat{t}_{k+1})^\top
    \end{bmatrix}^\top 
\end{align}
where $\mathbf{x}_{I_i}({}^{I_i}\hat{t}_{k+1})$ is the state of the $i$-th IMU at the estimated time ${}^{I_i}\hat{t}_{k+1}$ in its own clock.

\subsubsection{State Augmentation}
After propagating from timestep $k$ to $k+1$,
we have the (base) IMU state estimates at the {\em estimated} time ${}^{I_b}\hat{t}_{k+1}$ [see \eqref{eq:mimc-imu-state-prop}].
However, in order to update with all cameras' measurements at time step $k+1$, we actually need to express them as a function of the base IMU pose at the \textit{true} time ${}^{I_b}{t}_{k+1}$. 
To accomplish this,
we perform the following linearized stochastic cloning
in order to create an estimate of the base IMU at this true time~\cite{LiICRA2014}:
\begin{align}
    {}^G\mathbf{p}_{I}({}^{I_b}t_{k+1}) &\approx {}^G\mathbf{p}_{I}({}^{I_b}\hat{t}_{k+1}) + {}^G\mathbf{v}_{I}({}^{I_b}\hat{t}_{k+1}) {}^{C_b}\tilde{t}_{I_b} \label{eq::imu_temp_clone1}\\
    {}^{I({}^{I_b}t_{k+1})}_G\mathbf{R} &\approx \textrm{Exp}\left(-{}^{I_b}\bm\omega \left( {}^{I_b}\hat{t}_{k+1}\right) {}^{C_b}\tilde{t}_{I_b}\right){}^{I({}^{I_b}\hat{t}_{k+1})}_G\mathbf{R} \label{eq::imu_temp_clone2}
\end{align}
where 
${}^{I_b}\bm\omega\left( {}^{I_b}\hat{t}_{k+1}\right)$ is the true local angular velocity at time ${}^{I_b}\hat{t}_{k+1}$.
It is important to notice that,
because we can express the pose at the true time as a function of the pose at the estimated time (from propagation), 
as well as the time offset error (both of which are contained in our state vector~\eqref{eq:mimc-w-cal-state}), 
we can clone this new pose at the {\em true time} to include it into our state vector through stochastic cloning~\cite{Roumeliotis2002ICRAa}.

\subsubsection{FEJ Update with Multi-IMU Constraints}

We note that in order to enforce the multi-IMU constraints, we need to express each IMU at the same time. However, because we have uncertain time offsets, we have actually propagated each IMU to slightly \textit{different} times~\eqref{eq:mimc-imu-state-prop}.
As such,
we use a first-order approximation for the motion of each IMU 
and express the state of the $i$-th IMU at the exact time of the base IMU 
as a function of the state at the time we propagated to and the error in the time offset estimate. In particular, with abuse of notation, let ${}^{I_b}t = {}^{C_b}t_{k+1}+{}^{C_b}\hat{t}_{I_b}$ be the time the base IMU was propagated to. We approximate the $i$-th IMU pose at this true time as:
\begin{align}
    {}^G\mathbf{p}_{I_i} ({}^{I_i} t)
    &= {}^G\mathbf{p}_{I_i}({}^{I_b} t \!+\! {}^{I_b}\hat{t}_{I_i} \!+\! {}^{I_b}\tilde{t}_{I_i}) \notag \\
    &=  {}^G\mathbf{p}_{I_i}({}^{I_i}\hat{t} \!+\! {}^{I_b}\tilde{t}_{I_i}) \notag\\
    &\approx  {}^G\mathbf{p}_{I_i}({}^{I_i}\hat{t}) + {}^G\mathbf{v}_{I_i}({^{I_i}\hat{t}}) {}^{I_b}\tilde{t}_{I_i} \\
    {}^{I_i({}^{I_i} t)}_G\mathbf{R} &\approx \textrm{Exp}\left(-{}^{I_i}\bm \omega({}^{I_i}\hat{t}) {}^{I_b}\tilde{t}_{I_i}\right){}^{I_i({}^{I_i}\hat{t})}_G\mathbf{R} \label{eq:imu-calib2}
\end{align}
where 
 ${}^{I_i}\bm \omega({}^{I_i}\hat{t})$ is the angular velocity of the $i$-th IMU.
With that, we rewrite \eqref{eq:res_const_pos} in a residual form:
\begin{align}
    \scalemath{.95}{\mathbf{r}_{\theta_i}} &\scalemath{.80}{= 2\textbf{vec}\left( {}^{I_b({}^{I_b}{t})}_{G}{\bar{\mathbf{q}}} \otimes {}^{I_i(^{I_i}\hat{t})}_{G}{\bar{\mathbf{q}}}^{-1}\otimes \begin{bmatrix}
    \frac{1}{2}{{}^{I_i}\bm \omega}({}^{I_i}\hat{t}) {}^{I_b}\tilde{t}_{I_i} \\ 1
    \end{bmatrix}^{-1} \otimes {}^{I_b}_{I_i}\bar{\mathbf{q}}^{-1}\right)  \label{eq:res_const_pos_time0}
    }  \\
    \scalemath{.95}{\mathbf{r}_{p_i}} &\scalemath{.85}{=
    {}^G\mathbf{p}_{I_b}({}^{I_b}t) + {}^{I_b({}^{I_b}t)}_G\mathbf{R}^\top{}^{I_b}\mathbf{p}_{I_i} -
    {}^G\mathbf{p}_{I_i}({}^{I_i}\hat{t}) - {}^G\mathbf{v}_{I_i}({}^{I_i}\hat{t}) {}^{I_b}\tilde{t}_{I_i}  
    \label{eq:res_const_pos_time}}
\end{align}

It is important to note that the proposed MIMC-VINS leverages FEJ~\cite{Huang2008ISER,Huang2008ICRA} 
for all state variables (except calibration parameters) to perform EKF update with these multi-IMU measurements in order to improve estimation consistency,
which is different from  our previous work~\cite{Eckenhoff2019ICRAb}.
The FEJ-based observability constraints have been shown to greatly improve the MSCKF-based single-IMU/camera VINS~\cite{Li2013IJRR}
and we have also experimentally found that this FEJ treatment leads to large performance gains for the proposed MIMC-VINS estimator (see Section \ref{sec:sim_fej_compare}).

In particular, we apply FEJ to analytically compute the state-transition matrix as in~\cite{Hesch2013TRO}. 
As compared to the standard single-IMU case, to maintain consistency across the multi-IMU constraints, at every timestep
we use the first estimate for the base IMU pose along with the best estimate for the calibration  in order to form the linearization points for the auxiliary IMUs. 
Owing to the fact that we perform FEJ-based linearization in this described way,  the multi-IMU measurement Jacobians can be obtained as follows (see~\cite{Eckenhoff2019ICRAb}):
\begin{align}
    &\frac{\partial \mathbf{r}_{\theta_i}}{\partial {}^{I_b}\tilde{t}_{I_i}}  = -{}^{I_b}_{I_i}\hat{\mathbf{R}}\left(\bm \omega_{mi}-\hat{\mathbf{b}}_{gi}\right), ~~
 \frac{\partial \mathbf{r}_{\theta_i}}{\partial {}^{I_b}_{I_i}\tilde{\bm \theta}} = -\mathbf{I}, \notag\\
 &\frac{\partial \mathbf{r}_{\theta_i}}{\partial {}^{I_b({}^{I_b}\hat{t})}_G\tilde{\bm \theta}} = \mathbf{I}, ~~
\frac{\partial \mathbf{r}_{\theta_i}}{\partial {}^{I_i({}^{I_i}t)}_G\tilde{\bm \theta}} = -{}^{I_b}_{I_i}\hat{\mathbf{R}} \\
&\frac{\partial \mathbf{r}_{p_i}}{\partial {}^G\tilde{\mathbf{p}}_{I_i}({}^{I_i}\hat{t})} = -\mathbf{I}_3, ~
      {\frac{\partial \mathbf{r}_{p_i}}{\partial{}^G\tilde{\mathbf{p}}_{I_b}({}^{I_b}t)}} = {\mathbf{I}_3}, ~
      \frac{\partial \mathbf{r}_{p_i}}{\partial {}^{I_b}\tilde{t}_{I_i}} = {-{}^G\hat{\mathbf{v}}_{I_i}({}^{I_i}\hat{t})}, \notag\\
 &\frac{\partial \mathbf{r}_{p_i}}{\partial {}^{I_b({}^{I_b}t)}_G\tilde{\bm \theta}} = -{}^G_{I_b({}^{I_b}t)}\hat{\mathbf{R}}\lfloor{}^{I_b}\hat{\mathbf{p}}_{I_i}\times \rfloor, ~
      \frac{\mathbf{r}_{p_i}}{\partial {}^{I_b}\tilde{\mathbf{p}}_{I_i}} = {}^G_{I_b({}^{I_b}t)}\hat{\mathbf{R}} \label{eq:res_mi}
\end{align}

Note that as in the stochastic cloning of state augmentation, 
the value of the angular velocity used in the linearization [see \eqref{eq:imu-calib2}] 
comes from  the $i$-th IMU's gyro measurements (i.e. $ \bm \omega_{mi}({}^{I_i}\hat{t})- \hat{\mathbf{b}}_{gi}({}^{I_i}\hat{t})$). 
As before, errors in the linear and angular velocities are multiplied by ${}^{I_b}\tilde{t}_{I_i}$, and thus do not affect the measurement up to first order. 
It is also important to note that,
after update with the multi-IMU relative-pose constraints,
the state of the base IMU will still be at the \textit{prior} estimate at time  ${}^{I_b}\hat{t}_{k+1|{k}}$ (i.e., the time we last propagated to), while we will have an \textit{updated} estimate for the time offset ${}^{C_b} {\hat t}_{I_b}$.
To compensate for this, when we receive a new camera image at time ${}^{C_b}t_{k+2}$, 
we actually propagate the base IMU from ${}^{I_b}\hat{t}_{k+1|{k}}$ to ${}^{I_b}\hat{t}_{k+2|{{k+1}}} = {}^{C_b}t_{k+2}+ {}^{C_b}\hat{t}_{I_b}$
[see \eqref{eq:mimc-prop000}].
Similarly, the $i$-th non-base IMU will still be at the \textit{prior} estimate at time  ${}^{I_i}\hat{t}_{k+1|{k}}$, 
while we will have an \textit{updated} estimate for the time offset ${}^{I_b} {\hat t}_{I_i}$.
When we receive a new camera image at time ${}^{I_b}t_{k+2}$, 
propagation for each non-base IMU is  performed over the interval $[{}^{I_i}\hat{t}_{k+1|{k}}, {}^{I_i}\hat{t}_{k+2|{{k+1}}}]$
[see \eqref{eq:mimc-propiii}].

\subsubsection{FEJ Update with Multi-Camera Measurements} \label{sec:HOT_interp}

The linear interpolation in the barebones MIMC-VINS [see \eqref{eq:linear-intpl-1} and \eqref{eq:linear-intpl-2}]
relies on the assumption of \textit{linear} evolution in the orientation and position of the platform.
While this may be adequate in many scenarios, in the case of highly-dynamic motion, this model does not hold.
In particular, if performing update using this improper model, this constant-velocity motion assumption \textit{adds information} to the estimator that may be inconsistent, thereby destroying any performance gain from the addition sensors.

To address this issue, we generalize the linear interpolation to higher-order interpolation to capture more complex motion profiles seen in practice. Consider an image captured by the $i$-th camera, with reported timestamp ${}^{C_i}t$, which corresponds to time ${}^{C_b}t = {}^{C_i}t + {}^{C_i}t_{C_b}$ as expressed in the base camera clock. In order to process these measurements, we then need to be represent the pose at this true imaging time.
In particular, we fit a polynomial of degree $n$ in terms of time to a set of $n+1$ known poses 
(for simplicity of nations, we denote ${}^{C_i}t + {}^{C_i}t_{C_b} =: t$):
\begin{align} \label{eq:hot-intep00}
    {}^{I(t)}_G\mathbf{R} &= \textrm{Exp}\left(\sum_{i=1}^n \mathbf{a}_{\theta i} \Delta t^i \right){}^{I(t_0)}_G\mathbf{R} \\
    {}^G\mathbf{p}_{I}(t) &= {}^G\mathbf{p}_{I}(t_0) +     \sum_{i=1}^n\mathbf{a}_{p i} \Delta t^i
    \label{eq:hot-intep11}
\end{align}
where $t_0 := {}^{C_b}t_0$ represents the time of the oldest pose being fitted to as expressed in the base camera's clock and $\Delta t = t-t_0$. Note that due to representing the interpolated pose  as a polynomial function of the ``time change from $t_0$'', 
we trivially recover the pose at $t_0$ when plugging in $\Delta t = 0$.
Fitting this polynomial involves estimating the parameters $\mathbf{a}_{\theta i } $ and $\mathbf{a}_{p i}$. 
To do so, we require $n$ additional poses that, at each corresponding time, the polynomial should fit exactly (with corresponding time difference $\Delta t _k = t_k - t_0$:
\begin{align}
        {}^{I(t_k)}_G\mathbf{R} &= \textrm{Exp}\left(\sum_{i=1}^n \mathbf{a}_{\theta i} \Delta t_k^i \right){}^{I(t_0)}_G\mathbf{R} \\
        \Rightarrow \sum_{i=1}^n \mathbf{a}_{\theta i} \Delta t_k^i &= \textrm{Log}\left({}^{I(t_k)}_G\mathbf{R}{}^{I(t_0)}_G\mathbf{R}^\top\right) := \Delta \bm \phi_k \label{eq:log_diff}\\
    {}^G\mathbf{p}_{I}(t_k) &= {}^G\mathbf{p}_{I}(t_0) +
    \sum_{i=1}^n\mathbf{a}_{p i} \Delta t_k^i \\
    \Rightarrow \sum_{i=1}^n\mathbf{a}_{p i} \Delta t_k^i &= {}^G\mathbf{p}_{I}(t_k) -{}^G\mathbf{p}_{I}(t_0) := \Delta \mathbf{p}_k
    \label{eq:log_diff1}
\end{align}
These constraints can be stacked in order to solve for the stacked vectors of coefficients [see \eqref{eq:log_diff}]:
\begin{align}
    \underbrace{\begin{bmatrix} 
    \Delta \bm \phi_1 \\
    \Delta \bm \phi_2\\
    \vdots 
    \\ \Delta \bm \phi_n
    \end{bmatrix}}_{ \Delta \bm \phi} 
    &=  
    \underbrace{\begin{bmatrix}
    \Delta t_1 & \Delta t_1^2 & \cdots & \Delta t_1^n \\
    \Delta t_2 & \Delta t_2^2 & \cdots & \Delta t_2^n \\
    \vdots & \vdots & \ddots & \vdots \\
    \Delta t_n & \Delta t_n^2 & \cdots & \Delta t_n^n
    \end{bmatrix} }_{\mathbf{V}}
    \underbrace{
    \begin{bmatrix} 
    \mathbf{a}_{\theta 1} \\
    \mathbf{a}_{\theta 2} \\
    \vdots \\
    \mathbf{a}_{\theta n}
    \end{bmatrix} }_{\mathbf{a}_\theta} \\
    \mathbf{a}_\theta &= \mathbf{V}^{-1}\Delta \bm \phi
    \label{eq:hop-coeff1}
\end{align}
Proceeding analogously, we can obtain the coefficients for the position polynomial [see \eqref{eq:log_diff1}]:
\begin{align}
\mathbf{a}_p &= \mathbf{V}^{-1}\Delta \mathbf{p}
\label{eq:hop-coeff2}
\end{align}

These equations~\eqref{eq:hop-coeff1} and \eqref{eq:hop-coeff2} reveal that the interpolated pose and thus the corresponding measurements are dependent on the $n+1$ known poses which are in the state vector.
Note that this interpolated pose is additionally a function of the unknown time offset, and thus we take the Jacobian with respect to this parameter and estimate it \textit{online}. These measurement Jacobians can be computed as shown in Appendix~\ref{appendix::HOI}.

In order to implement this in the proposed MIMC-VINS, for each auxiliary camera measurement we query the poses which bound estimated measurement time to form the polynomial for fitting.
We attempt to make the segment that the measurement falls in to be in the ``middle'' of the set of poses.
For example, when forming an odd order polynomial, we try to ensure that we have $(n+1)/2$ poses earlier and newer than the measurement.
If we do not have enough poses earlier or later, we simply grab the earliest/latest $n+1$ poses.

It is important to note that  we also employ the FEJ methodology 
when computing measurement Jacobians involved with the above interpolation in order to improve estimation consistency,
 but use the similar methods in~\cite{Li2014,Paul2018CVPR} when computing the residual. 
In particular, we use the saved IMU readings to propagate from the closest neighbor clone estimate to compute ${}^{G}\hat{\mathbf{p}}_{I}(\hat{t})$, as this offers a more accurate estimate for the interpolated pose.

\subsection{FEJ Update with SLAM Features}

In order to further improve estimation accuracy, as in \cite{Li2013RSS}, 
we selectively keep certain visual features (aka SLAM features) -- which can be reliably tracked beyond the sliding window and are denoted by $\mathbf{x}_m=[{}^G\mathbf{p}_{f1}^\top~\cdots~{}^G\mathbf{p}_{fk}^\top]^\top$ -- 
in the state vector [see \eqref{eq:mimc-w-cal-state}]  till they get lost and are then marginalized.
Importantly, when performing EKF update with the measurements of these SLAM features,
we again enforce the FEJ-based observability constraints in computing these measurement Jacobians in order to further improve estimation consistency and thus accuracy~\cite{Huang2008ICRA}.
Note that we use the process of delayed initialization to add these features to our state~\cite{Li2014}.

\subsection{Extending to Rolling-Shutter Cameras}

We have thus far assumed global-shutter cameras wherein all measured pixel intensities for  single image correspond to the same instance in time. 
However, low-cost cameras often utilize a \textit{rolling shutter (RS)}, wherein each row of the image is captured sequentially. 
This may lead to large errors in the estimation if this RS effect is not taken into account~\cite{Guo2014RSS}. 
To address this issue, we also perform online calibration of the RS readout time.
Specifically, let the total readout time of the $i$-th camera be $t_{ri}$, and $t$ denote the time that the pixel was captured. 
The RS measurement function for a pixel captured in the $m$-th row (out of $M$ in total) of the $i$-th camera is given by [see \eqref{eq:feature_in_cam0}]:
\begin{align}
    \mathbf{z}_{i}(t) &= \mathbf{w}_i\left(\bm \Pi \left({}^{C_i\left(t\right)}\mathbf{p}_f\right),~ \bm \zeta_i \right)+ \mathbf{n}_{zi}(t)
     \label{eq::mc_update}\\
    t &= {}^{C_i}t + {}^{C_i}t_{C_b}+ \frac{m}{M}t_{ri} \label{eq::rs_time}
\end{align}
where ${}^{C_i}t$ is the nominal imaging start time in the $i$-th camera clock and ${}^{C_i}t_{C_b}$ is the time offset as previously discussed. 
Note that one may define this nominal imaging time in different ways (e.g. start of image, end of image, middle of image), 
which simply requires minor modification of \eqref{eq::rs_time} \cite{Li2014IJRRb,Guo2014RSS}.

The RS effect simply adds a further offset into the time the measurement was captured. 
These measurements can be seamlessly incorporated into our MIMC-VINS estimator with one slight modification; 
\textit{all} measurements, even the base camera,
must be expressed using the proposed high-order polynomial interpolation.
In addition, we will have an extra Jacobian with respect to the unknown RS readout time when updating with camera measurements:
\begin{align}
    \frac{\partial \mathbf{z}_i}{\partial \tilde{t}_{ri}} = \frac{m}{M}\frac{\partial \mathbf{z}_i}{\partial {}^{C_i}\tilde{t}_{C_b}}
\end{align}

\subsection{Resilience to Sensor Failures}
\label{sec:sensor-fail}

When navigating in challenging environments,
robustness to sensor failures is key to persistent localization.
The proposed MIMC-VINS is resilient to up to $N$ IMU and $M$ camera sensor failures.
While the failure of cameras might not be catastrophic as failed cameras simply do not provide any measurements for update, 
the failure of the IMU prevents VINS from performing state estimation (as the IMU is the backbone of propagation).
In the case that the base camera fails, we simply perform stochastic cloning at an arbitrarily chosen frequency.

We consider the more challenging scenario where the base IMU sensor fails, 
as when a non-base IMU sensor fails it is trivial to marginalize its navigation state.
During the base failure, we ``promote'' an auxiliary sensor to be the new base (which is denoted  as the $n$-th IMU).
We then transform the quantities in our state to be expressed with respect to this new base.
Specifically, each IMU clone refers to the base IMU sensor frame at the true imaging time, $t_j$, and we can write the following transformations for each:
\begin{align} \label{eq:mimc-change-imu-base1}
    {}^{I_n(t_j)}_G\mathbf{R} &= {}^{I_b}_{I_n}\mathbf{R}^{\top} {}^{I_b(t_j)}_G\mathbf{R} \\
    {}^{G}\mathbf{p}_{I_n}(t_j) &= {}^{G}\mathbf{p}_{I_b}(t_j)+ {}^{I_b(t_j)}_G\mathbf{R}^{\top}{}^{I_b}\mathbf{p}_{I_n}
    \label{eq:mimc-change-imu-base2}
\end{align}
Additionally we can propagate the IMU-IMU and camera-IMU transforms:
\begin{align}
    {}^{I_n}_{I_i}\mathbf{R} &= {}^{I_n}_{I_b}\mathbf{R}^\top {}^{I_b}_{I_i}\mathbf{R}, ~~~
    {}^{I_n}\mathbf{p}_{I_i} = {}^{I_n}_{I_b}\mathbf{R}\left({}^{I_b}\mathbf{p}_{I_i}- {}^{I_b}\mathbf{p}_{I_n} \right) \\
     {}^{C_k}_{I_n}\mathbf{R} &= {}^{C_k}_{I_b}\mathbf{R} {}^{I_b}_{I_n}\mathbf{R}^{\top} , ~
    {}^{C_k}\mathbf{p}_{I_n} = {}^{C_k}\mathbf{p}_{I_b} + {}^{C_k}_{I_b}\mathbf{R}{}^{I_b}\mathbf{p}_{I_n} 
\end{align}
Moreover, the relationship between any clock, the base clock, and the new base clock takes the form:
\begin{align}
{}^{I_b}t+{}^{I_b}t_{I_i}= {}^{I_n}t+{}^{I_n}t_{I_i} ~\Rightarrow~ {}^{I_n}t_{I_i} = {}^{I_b}t_{I_i}-{}^{I_b}t_{I_n}
\end{align}
Using these constraints, 
we can modify the estimates such that the $n$-th IMU serves as the new base, through a proper mean and covariance propagation.
It is important to note that this procedure can be triggered at any point, such as when base sensor failure is detected, thus allowing for {\em continuous} and {\em uninterrupted} estimation.

\subsection{Remarks and Summary}

At this point, we have explained in detail the key advanced features of the proposed MIMC-VINS with online sensor calibration, improved resilience, and consistency,
which enables the plug-and-play functionality and accurate estimation performance.
We here share some important remarks: 
\begin{itemize}
\item Online sensor calibration of both intrinsics and extrinsics is important, 
and not only enables the plug-and-play functionality (and thus lowers the technology barriers for end users) but often improves the estimation performance.
We also note that we chose to not perform online IMU intrinsic calibration due to the high number of degenerate motions which can adversely affect accuracy \cite{Yang2020RSS}.
\item The FEJ-based methodology is essential for improving estimation consistency/accuracy
and has been utilized in multiple new ways; 
in addition to being used in processing SLAM feature measurements in analogy to EKF-SLAM~\cite{Huang2008ISER}, 
it is employed in both multi-IMU constraints and high-order interpolated multi-camera measurements.
\item It is often required in autonomous robots to provide continuous uninterrupted localization solutions, for example, in order to support motion planning,
even when some sensor fails during operation. 
The proposed MIMC-VINS offers such resilience by seamlessly switching the failed sensor states to the healthy ones without interrupting localization continuity,
and thus greatly improves system reliability and extends mission duration.
\end{itemize}
The main steps of the proposed MIMC-VINS have been summarized and outlined in Algorithm \ref{alg:MIMC-VINS}.

\begin{algorithm}
\caption{Versatile and Resilient MIMC-VINS Estimation}

\begin{flushleft}
\textbf{Input}: $N+1$ IMUs and $M+1$ cameras' measurements.\\
\textbf{Output:} Inertial navigation states and intrinsic/extrinsic calibration parameters, as well as their error covariance.
\end{flushleft}

\begin{algorithmic}[1]

\STATE {\tt Initialization:} Initialize state estimate and covariance with first few inertial/visual measurements.
Assume the first IMU and camera are the respective base sensors.

\LOOP

\STATE {\tt Propagation:}
When all IMU measurements available in the consecutive base imaging time interval $[{}^{I_i}{\hat t}_k, {}^{I_i}{\hat t}_{k+1}]$ ($i=0,\cdots,N$):

\IF{IMU sensor failure detected}
\IF{Failed IMU is the base}
\STATE Switch to an arbitrarily chosen new IMU base [see \eqref{eq:mimc-change-imu-base1} and \eqref{eq:mimc-change-imu-base2}].
\ENDIF
\STATE Marginalize the failed IMU state (by removing the corresponding state estimate and covariance blocks).
\ENDIF

\STATE Propagate each IMU state from time ${}^{I_i}\hat{t}_{k}$ to  
${}^{I_i}\hat{t}_{k+1} = {}^{C_b}t_{k+1} +{}^{C_b}\hat{t}_{I_b} + {}^{I_b}\hat{t}_{I_i}$ 
[see \eqref{eq:mimc-prop000} and \eqref{eq:mimc-propiii}].

\STATE {\tt State Augmentation:} 
Stochastic cloning of the base IMU pose at the true time ${}^{C_b}t_{k+1}$ [see \eqref{eq::imu_temp_clone1} and \eqref{eq::imu_temp_clone2}].

\STATE {\tt Inertial Update:} 
EKF update with multi-IMU relative-pose constraints [see \eqref{eq:res_const_pos_time0} and \eqref{eq:res_const_pos_time}],
where FEJ is imposed in computing their Jacobians.

\STATE {\tt Visual Update:} 
When camera images available at the base imaging time ${}^{C_b}t_{k+1}$:

\IF{Camera failure detected}
\IF{Failed camera is the base}
\STATE Switch to arbitrarily cloning at a fixed frequency.
\ENDIF 
\STATE Skip this faulty camera. 
\ENDIF

\STATE Perform KLT-based visual feature tracking for each healthy camera, which can be parallelized if possible (see Section~\ref{sec:parallel-tracking}). 

\STATE Perform FEJ-MSCKF update with multi-camera (RS) measurements [see~\eqref{eq::mc_update}],
where the high-order polynomial interpolation is employed [see \eqref{eq:hot-intep00} and \eqref{eq:hot-intep11}],
and FEJ is imposed in processing all feature observations.

\ENDLOOP
\end{algorithmic}
\label{alg:MIMC-VINS}
\end{algorithm}

 \section{Simulation Results} \label{sec:sim}

The proposed MIMC-VINS estimator is developed within our recently open-sourced OpenVINS~\cite{Geneva2019IROSws,Geneva2020ICRA}. This code base provides a visual-inertial MSCKF-based estimation framework including a set of simulation and evaluations tools, and its basic single-camera/IMU VINS has been shown to outperform existing open-sourced state-of-the-art methods.
To fully understand the effects of sensing properties on estimation performance and validate the proposed algorithm, 
in this section we perform extensive simulation tests on the synthetic data generated by our general visual-inertial simulator.
We believe this is important because simulations can help crystallize and validate our design choices on the scenarios of interest that may be hard to encounter in experiments,
establish the limitations of the state of the art, and demonstrate issues to be addressed.

\subsection{MIMC-VINS Simulator}
\label{sec:simulator}

The new MIMC-VINS simulator is generalized from the OpenVINS simulator~\cite{Geneva2020ICRA}
to be able to encompass an arbitrary number of IMUs and cameras and  simulate the proposed measurement models.
In the following we explain in detail the key designs of this general simulator,
while the exact parameters used in the simulations presented below are given in Table \ref{tab:sim_params}.

\begin{table}
\centering
\caption{
Simulation parameters and prior single standard deviations that perturbations of measurements and initial states were drawn from.
Note that if values were applied uniformly to all sensors only a single value is reported in this table.
}
\label{tab:sim_params}
\begin{adjustbox}{width=\columnwidth,center}
\begin{tabular}{ccccc} \toprule
\textbf{Parameter} & \textbf{Value} & \textbf{Parameter} & \textbf{Value} \\ \midrule
Cam Freq. (hz) & 10,11,13,23,18,22 & IMU Freq. (hz) & 400 \\
Num. Feats Per Camera & 25 & Num. SLAM feats & 0 \\
Num. Base Cam. Clones & 10 & Feat. Rep. & GLOBAL \\
Gyro. Bias Turn-on Prior & 0.01 & Accel. Bias Turn-on Prior & 0.01 \\
Gyro. White Noise & 1.6968e-04 & Gyro. Rand. Walk & 1.9393e-05 \\
Accel. White Noise & 2.0000e-3 & Accel. Rand. Walk & 3.0000e-3 \\
Pixel Noise & 1 & IMU Constraint Noise & 0.005 \\
Prior Calib Ori. (rad) & 0.017 & Prior Calib Pos. (m) & 0.01 \\
Prior Calib Toff (sec) & 0.01 & Prior Readout (sec) & 0.003 \\
Prior Cam Proj & 1.0 & Prior Cam Dist & 0.01 \\
\bottomrule
\end{tabular}
\end{adjustbox}
\end{table}

\subsubsection{B-Spline Trajectory Representation}

\begin{figure}
\centering
\includegraphics[trim={0 0 0 1.8cm},clip,width=.75\columnwidth]{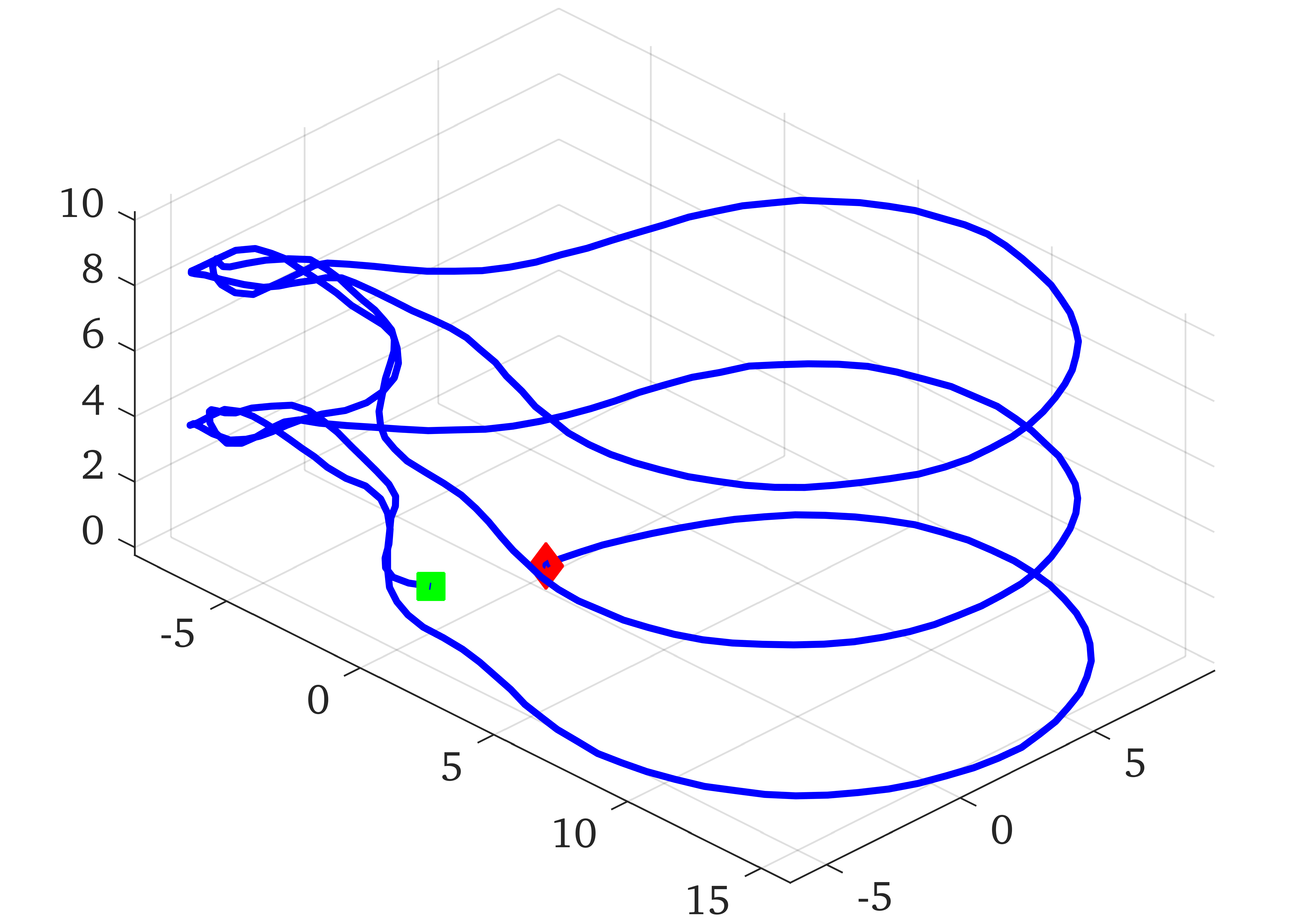}
\includegraphics[width=.70\columnwidth]{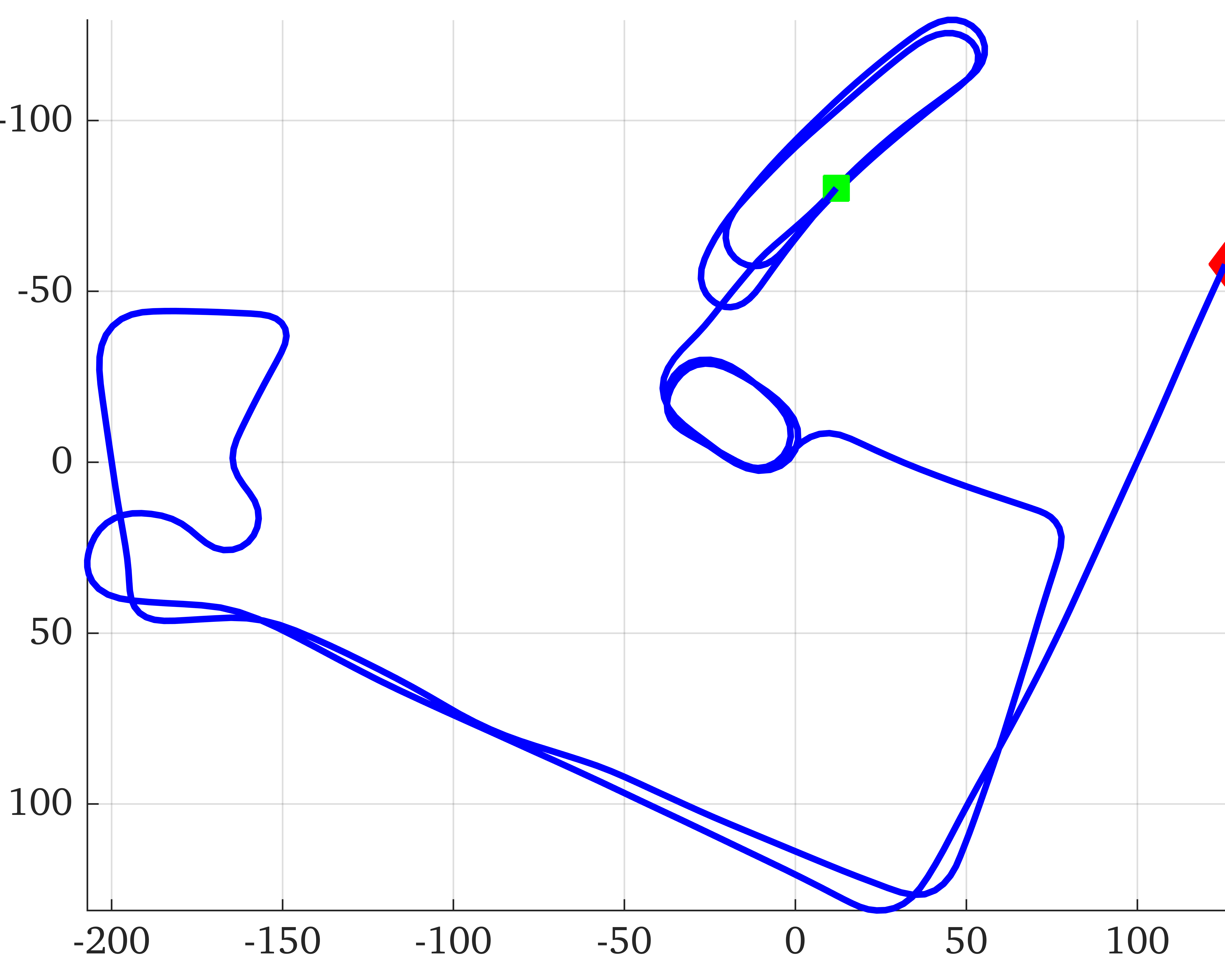}
\includegraphics[width=.75\columnwidth]{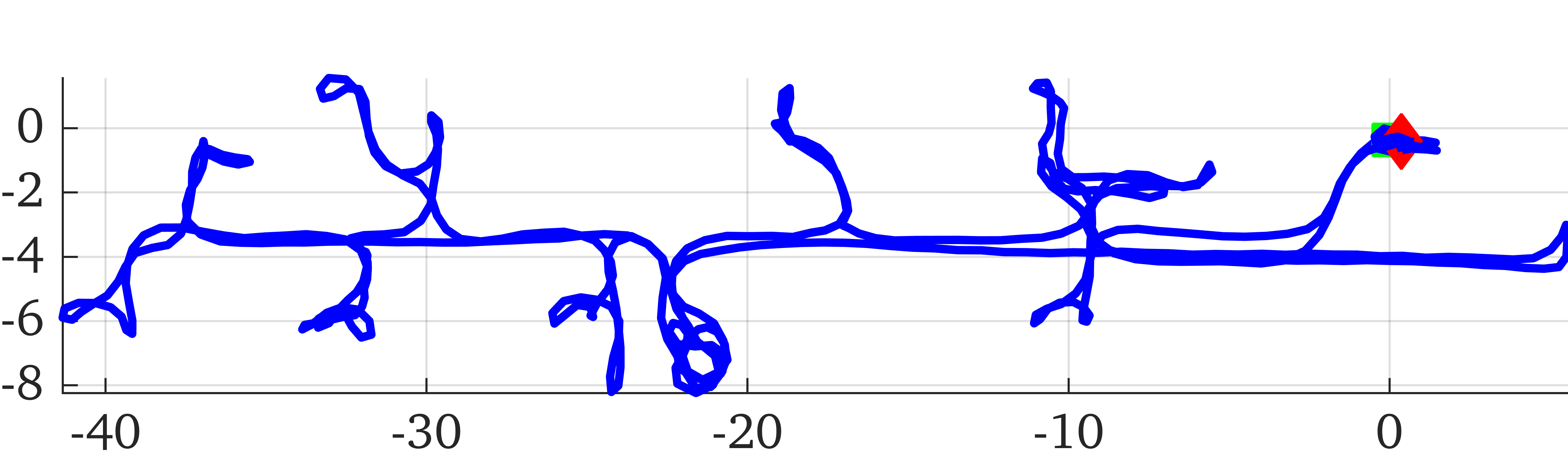}
\caption{
Simulated trajectories, axes are in units of meters.
Going from top to bottom: Gore, Outdoor, and Tum.
Green square denotes the start and red diamond denotes the end.
}
\label{fig:mimc_simulation_trajectories}
\end{figure}

At the center of our simulator is a  $\mathbb{SE}(3)$ B-spline representing a continuous-time trajectory of 6DOF poses,
which allows for the calculation of the pose, velocity, and accelerations at any given timestep along the trajectory.
For the simulation results presented in this section, the input is a pose trajectory file which is uniformly sampled and a cubic B-spline is fitted to.
The three simulated but realistic trajectories used to evaluate the proposed approach are depicted in Figure~\ref{fig:mimc_simulation_trajectories}:
(i) The first simulated dataset is called ``Gore'' and is based on a 240 meter dataset collected in the University of Delaware's Gore Hall in which three floors are traversed;
(ii) The second dataset is called ``Outdoor'' and was collected by a VINS system mounted on a car which travels along three levels of a parking garage before exiting and driving along the road, and has total length of approximately 1800 meters;
(iii) The third scenario is highly dynamic and based on the ``Tum Corridor 1'' dataset from the TUM VI benchmark datasets~\cite{schubert2018vidataset} which we simply term ``Tum'' and is approximately 295 meters.

\subsubsection{Inertial Measurements}

By computing the time derivatives of the B-spline,
we can obtain the true angular velocity and linear acceleration of a single IMU (i.e., the base simulated IMU $I_b$) that is attached to the trajectory.
Specifically, we leverage the B-spline to compute the linear velocity ${}^G\dot{\mathbf{p}}_{I_b} = {}^G{\mathbf{v}}_{I_b}$ and acceleration
${}^G\ddot{\mathbf{p}}_{I_b} = {}^G{\mathbf{a}}_{I_b}$,
while the angular rate ${}^{I_b}\bm \omega_{I_b}$ and angular acceleration ${}^{I_b}\bm \alpha_{I_b}$ are given by:
\begin{align}
{}^G_{I_b}\dot{\mathbf{R}} &= {}^G_{I_b}{\mathbf{R}}\lfloor {}^{I_b}\bm \omega_{I_b} \times \rfloor \\
\Rightarrow {}^{I_b}\bm \omega_{I_b} &= \left({}^G_{I_b}{\mathbf{R}}^{\top}{}^G_{I_b}\dot{\mathbf{R}}\right)^{\vee} \\
{}^G_{I_b}\ddot{\mathbf{R}} &= {}^G_{I_b}\dot{\mathbf{R}}\lfloor {}^{I_b}\bm \omega_{I_b} \times \rfloor + {}^G_{I_b}{\mathbf{R}}\lfloor {}^{I_b}\bm \alpha_{I_b} \times \rfloor \\
\Rightarrow {}^{I_b}\bm \alpha_{I_b} &= \left({}^G_{I_b}{\mathbf{R}}^{\top}\left({}^G_{I_b}\ddot{\mathbf{R}} -{}^G_{I_b}\dot{\mathbf{R}}\lfloor {}^{I_b}\bm \omega_{I_b} \times \rfloor \right)\right)^{\vee}
\end{align}
where $(\cdot)^{\vee}$ extracts the vector of a skew symmetric matrix.
With these kinematics, the linear acceleration and angular velocity of a non-base IMU $I_i$
can be computed based on the rigid-body constraints between the two IMUs~\cite{Rehder2016}:
\begin{align}
    {}^G\mathbf{a}_{I_i} &=  {}^G\mathbf{a}_{I_b} + {}^{G}_{I_b}\mathbf{R}\left(\lfloor {}^{I_b}\bm \omega_{I_b} \times \rfloor \lfloor{}^{I_b}\bm \omega_{I_b} \times \rfloor +\lfloor {}^{I_b}\bm \alpha_{I_b} \times \rfloor \right){}^{I_b}\mathbf{p}_{I_i} \\
    {}^{I_i}\bm \omega_{I_i} &= {}^{I_i}_{I_b}\mathbf{R}{}^{I_b}\bm \omega_{I_b}
\end{align}

Based on the above kinematic equations,
we generate the synthetic true inertial measurements for each IMU at any given time with any given extrinsic transformation,
which are then corrupted using the random walk biases and white noise to simulate realistic IMU readings.
Note that when simulating with non-zero time offset between the the base and non-base IMUs, the true timestamp is changed based on the true time offset between the two.

\subsubsection{Visual Bearing Measurements}

After generating the B-spline-based trajectory, 
static environmental features are incremented along the trajectory and if the number of projected features fall below the desired feature count threshold, random feature bearing rays are generated from each camera and assigned random depths.
These  features are then appended to the environmental map and projected into future frames.

In the case of a global shutter camera, visual feature measurements are simulated by projecting the corresponding environmental features into the camera frame using the true camera parameters.
These true raw pixel coordinates are then corrupted using their corresponding white noise.

In the case of a rolling shutter camera,
feature measurements were generated with an additional modification to the feature projection procedure \cite{Li2014IJRRb}.
In particular, due to the RS effect one cannot simply utilize the groundtruth feature positions and camera poses to project onto the image plane, as each row was captured at a different time.
Therefore, we start with the nominal image time $t_0$, and perform standard global shutter projection for a given feature.
This projection yields a row $m_0$ that the feature projected to.
With this row, we then compute a \textit{new} imaging time based on the RS effect [see~\eqref{eq::rs_time}]:
\begin{align}
    t_1 = t_0 + \frac{m_0}{M}t_{ri}
\end{align}
Using this new time, we recompute the projection using the camera pose at $t_1$, and iterate this process till $t_1$ converges (which typically requires 2-3 iterations).
This ensures that for a given feature, it is captured at the pose time corresponding to the image measurement time plus the rolling shutter readout time.
These true raw pixel coordinates are then corrupted by white noise. Lastly, in order to mimic poor prior calibration, 
we perturbed each of the calibration parameters by sampling from their prior distributions (see Table \ref{tab:sim_params} for all parameters used in the simulations).

Based on this general visual-inertial simulator, in the following, 
we first perform numerical studies of the impact of camera and inertial sensors on estimation performance, 
and then present the full MIMC-VINS running in simulations.

\subsection{Multi-Camera Simulations}

\subsubsection{Effect of Adding Cameras}

To investigate the benefit of adding extra cameras into the multi-camera visual-inertial system,
we first perform 30 Monte-Carlo simulations for each dataset and each number of cameras.
For these runs we report both the Absolute Trajectory Error (ATE), which is the average RMSE across the dataset, and the Relative Pose Error (RPE) \cite{Zhang2018IROS}. For these experiments, each camera was allowed to track 25 features to mimic the effects of measurement depletion in a given viewing direction.
The results are shown in Tables \ref{table::ate_MC} and \ref{table::rpe_MC},
which clearly show that adding more cameras  tends to improve the estimation performance in term of both ATE and RPE.
In particular, the six-camera configuration has errors almost half of the monocular VINS,
which greatly validates our desire to add more cameras into the system for building more accurate VINS, even if the system suffers poor prior calibration.

\begin{table*}
\centering
\caption{ATE in degrees/meters when performing online calibration with multiple cameras.}
\begin{tabular}{ccccc} \toprule 
\textbf{Number of Cameras} & \textbf{Outdoor} & \textbf{Tum} & \textbf{Gore} & \textbf{Average} \\\midrule
1 & 0.277 / 1.480 & 1.173 / 0.828 & 1.824 / 0.639 & 1.091 / 0.982 \\
2 & 0.250 / 1.133 & 0.891 / 0.444 & 1.057 / 0.450 & 0.733 / 0.676 \\
3 & 0.231 / 1.017 & 0.649 / 0.374 & 0.752 / 0.375 & 0.544 / 0.589 \\
4 & 0.228 / 0.945 & 0.468 / 0.270 & 0.557 / 0.297 & 0.418 / 0.504 \\
5 & 0.214 / 0.852 & 0.451 / 0.240 & 0.518 / 0.252 & 0.394 / 0.448 \\
6 & \textbf{0.205} / \textbf{0.816} & \textbf{0.329} / \textbf{0.221} & \textbf{0.460} / \textbf{0.238} & \textbf{0.331} / \textbf{0.425} \\
 \bottomrule
\end{tabular}
\label{table::ate_MC}
\end{table*}

\begin{table*}
\centering
\caption{RPE in degrees/meters for the simulated datasets using different numbers of cameras.}
\begin{tabular}{ccccccc} \toprule 
\textbf{Num. Cameras}  & \textbf{8m} & \textbf{16m} & \textbf{24m} & \textbf{32m} & \textbf{40m} & \textbf{48m} \\\midrule
1 & 0.117 / 0.138 & 0.156 / 0.192 & 0.181 / 0.237 & 0.204 / 0.277 & 0.218 / 0.309 & 0.233 / 0.342 \\
2 & 0.097 / 0.092 & 0.131 / 0.130 & 0.149 / 0.161 & 0.166 / 0.187 & 0.179 / 0.208 & 0.192 / 0.227 \\
3 & 0.083 / 0.077 & 0.111 / 0.110 & 0.127 / 0.137 & 0.142 / 0.159 & 0.151 / 0.179 & 0.159 / 0.195 \\
4 & 0.071 / 0.064 & 0.096 / 0.090 & 0.109 / 0.112 & 0.118 / 0.131 & 0.128 / 0.148 & 0.137 / 0.162 \\
5 & 0.066 / 0.057 & 0.087 / 0.080 & 0.100 / 0.099 & 0.109 / 0.116 & 0.116 / 0.131 & 0.126 / 0.145 \\
6 & \textbf{0.062} / \textbf{0.051} & \textbf{0.082} / \textbf{0.073} & \textbf{0.094} / \textbf{0.091} & \textbf{0.103} / \textbf{0.108} & \textbf{0.110} / \textbf{0.123} & \textbf{0.117} / \textbf{0.136} \\
 \bottomrule
\end{tabular}
\label{table::rpe_MC}
\end{table*}

\subsubsection{Implication of Multi-Camera Features}

In general, as more cameras are added and provide a larger field of view, the total number of tracked features would increase
(e.g. if one camera tracks 50 features in total, then for two cameras 100 features may be tracked).
The increased features can provide more information to the VINS estimator, 
thus allowing to further reduce uncertainty and increase accuracy.
One may ask  what happens if the total number of tracked features is simply kept constant as we add more cameras
(e.g. one camera extracts 100 features, two cameras each extract 50).
While a more intelligent and adaptive feature selection strategy can be devised, 
in this simulation (where features are roughly uniformly distributed in the environment),
we proportionally reduce the number of features that each camera tracks if adding more cameras in order to keep the same total number of tracked features.

The results in Table \ref{table::feat_max_tracks} 
show that a single camera which extracts 150 features is able to achieve the same level of accuracy as a three camera system which extracts the same number of features.
However,  this makes sense in this particular test as there is approximately the same amount of information  added into the system 
due to the fact that the features can be uniformly tracked by different cameras, 
which typically is not the case in practice.
One of the key benefits to including multiple cameras even if the amount of tracked features is not increased is to improve the robustness to viewpoint failures.
For example, if one camera faces a textureless wall, it would be unable to extract features and provide motion information, 
while in a multi-camera system the other cameras provide robustness by viewing the environment from different directions.

\begin{table}[t]
\centering
\caption{
ATE in degrees/meters for adding more features and a constant total number as tested on the Tum dataset.
}
\begin{tabular}{cccc} \toprule
\textbf{Num. Camera} & \textbf{Feat. Per Cam.} & \textbf{Feat. Total} & \textbf{ATE} \\\midrule
1 & 150 & 150 & 0.507 / 0.362 \\
2 & 75 & 150 & 0.589 / 0.298 \\
3 & 50 & 150 & 0.564 / 0.304  \\
\midrule
1 & 50 & 50 & 0.950 / 0.744  \\
2 & 50 & 100 & 0.747 / 0.354 \\
3 & 50 & 150 & 0.564 / 0.304 \\
\bottomrule
\end{tabular}
\label{table::feat_max_tracks}
\end{table}

\subsubsection{Choice of Interpolation Order}

To understand  how the choice of interpolation order affects performance,
we tested on the most dynamic of the three datasets, Tum, and used an extremely low frequency of 5 Hz for the base camera.
Six cameras were utilized in total, and calibration was performed online.
The RPE and ATE results are shown in Tables \ref{table::ate_tum_interp_order} and \ref{table::multi_cam_interp}.
Interestingly, while accuracy (in terms of both metrics) is improved when moving from order one to order three, adding additional complexity to the polynomials tends to cause a \textit{decrease} in accuracy. We conjecture that there may be overfitting in the polynomial regression as the system ``interprets'' small errors in the estimates as higher-order motion. This is especially problematic as we leverage FEJ to compute Jacobians, and thus the system attempts to fit a polynomial to a series of suboptimal, \textit{initial} estimates, rather than a smoothed window of the current \textit{full} estimates. Overall, we found that order-three polynomials offer the best performance, albeit by a small margin. 
We also note that one could alternatively utilize different orders for the position and orientation interpolation independently.

\begin{table}[t]
\centering
\caption{ATE in degrees/meters for the Tum dataset when using different interpolation orders for six cameras.}
\begin{tabular}{cc} \toprule
\textbf{Interp. Order} & \textbf{ATE} \\\midrule
1 & 0.274 / 0.180 \\
3 & \textbf{0.270} / \textbf{0.179} \\
5 & 0.273 / 0.180 \\
7 & 0.302 / 0.181  \\
 \bottomrule
\end{tabular}
\label{table::ate_tum_interp_order}
\end{table}
\begin{table*}
\centering
\caption{RPE  in degrees/meters for the simulated datasets using different interpolation orders.}
\begin{tabular}{ccccccc} \toprule
\textbf{Interp. Order}  & \textbf{8m} & \textbf{16m} & \textbf{24m} & \textbf{32m} & \textbf{40m} & \textbf{48m} \\\midrule
1 & 0.070 / 0.027 & 0.084 / 0.036 & 0.092 / 0.043 & 0.099 / 0.050 & 0.105 / 0.056 & 0.113 / 0.064 \\
3 & \textbf{0.063} / \textbf{0.027} & \textbf{0.077} / \textbf{0.035} & \textbf{0.085} / \textbf{0.042} & \textbf{0.093} / \textbf{0.049} & \textbf{0.099} / \textbf{0.054} & \textbf{0.107} / \textbf{0.062} \\
5 & 0.068 / 0.027 & 0.082 / 0.036 & 0.090 / 0.043 & 0.097 / 0.050 & 0.103 / 0.055 & 0.111 / 0.063 \\
7 & 0.075 / 0.027 & 0.088 / 0.036 & 0.097 / 0.043 & 0.104 / 0.050 & 0.111 / 0.056 & 0.118 / 0.064 \\
\bottomrule
\label{table::multi_cam_interp}
\end{tabular}
\end{table*}

\subsection{Multi-IMU Simulations}

\subsubsection{Effect of Adding IMUs}

We now numerically evaluate the accuracy gains  when adding additional IMUs into the system.
In this test, we initialized the velocities and biases of each IMU, along with the pose of the base IMU, as the groundtruth.
We then used the uncertain imperfect calibration parameters to clone the pose estimates  of each auxiliary IMU.
For these experiments, six noisy IMUs were simulated with varying spatial extrinsics.
We performed 30 Monte-Carlo simulations of the proposed estimator for each number of IMU used. 
The RPE and ATE results are shown in Tables \ref{table::ate_MI} and \ref{table::rpe_MI},
which clearly reveal that  adding more IMUs always improves both the ATE and the RPE for every segment tested.
Overall, these results confirm that adding IMUs can dramatically improve estimation even with poor prior calibration
and  validate our desire to add additional inertial sensors into the visual-inertial system.

\begin{table*}
\centering
\caption{ATE  in degrees/meters when performing online calibration with multiple IMUs.}
\begin{tabular}{ccccc} \toprule
\textbf{Num. IMU} & \textbf{Outdoor} & \textbf{Tum} & \textbf{Gore} & \textbf{Average} \\\midrule
1 & 0.277 / 1.480 & 1.173 / 0.829 & 1.824 / 0.640 & 1.091 / 0.983 \\
2 & 0.217 / 1.197 & 1.112 / 0.713 & 1.548 / 0.515 & 0.959 / 0.808 \\
3 & 0.190 / 1.078 & 1.018 / 0.655 & 1.418 / 0.465 & 0.875 / 0.733 \\
4 & 0.180 / 1.014 & 0.773 / 0.593 & 1.332 / 0.436 & 0.762 / 0.681 \\
5 & 0.176 / 0.986 & 0.703 / 0.587 & 1.222 / 0.415 & 0.700 / 0.662 \\
6 & \textbf{0.166} / \textbf{0.922} & \textbf{0.663} / \textbf{0.571} & \textbf{1.143} / \textbf{0.398} & \textbf{0.657} / \textbf{0.630} 
 \\\bottomrule
\end{tabular}
\label{table::ate_MI}
\end{table*}

\begin{table*}
\centering
\caption{RPE  in degrees/meters for the large-scale simulated dataset using different numbers of IMUs.}
\begin{tabular}{ccccccc}\toprule
\textbf{Num. IMU}  & \textbf{8m} & \textbf{16m} & \textbf{24m} & \textbf{32m} & \textbf{40m} & \textbf{48m} \\\midrule
1 & 0.117 / 0.138 & 0.156 / 0.192 & 0.181 / 0.237 & 0.204 / 0.277 & 0.218 / 0.309 & 0.233 / 0.342 \\
2 & 0.098 / 0.126 & 0.137 / 0.176 & 0.160 / 0.217 & 0.184 / 0.251 & 0.194 / 0.280 & 0.203 / 0.306 \\
3 & 0.091 / 0.120 & 0.123 / 0.167 & 0.144 / 0.206 & 0.163 / 0.236 & 0.177 / 0.264 & 0.182 / 0.289 \\
4 & 0.081 / 0.115 & 0.111 / 0.160 & 0.131 / 0.196 & 0.147 / 0.225 & 0.161 / 0.251 & 0.168 / 0.273 \\
5 & 0.078 / 0.112 & 0.106 / 0.156 & 0.126 / 0.191 & 0.140 / 0.219 & 0.153 / 0.244 & 0.162 / 0.264 \\
6 & \textbf{0.075} / \textbf{0.110} & \textbf{0.101} / \textbf{0.152} & \textbf{0.121} / \textbf{0.185} & \textbf{0.134} / \textbf{0.213} & \textbf{0.145} / \textbf{0.237} & \textbf{0.153} / \textbf{0.257} 
 \\\bottomrule
\end{tabular}
\label{table::rpe_MI}
\end{table*}

\subsection{Complete MIMC-VINS Simulations}

\subsubsection{Calibration Consistency and Convergence}

\begin{figure*}
\centering
\subfloat{\includegraphics[width=0.32\textwidth]{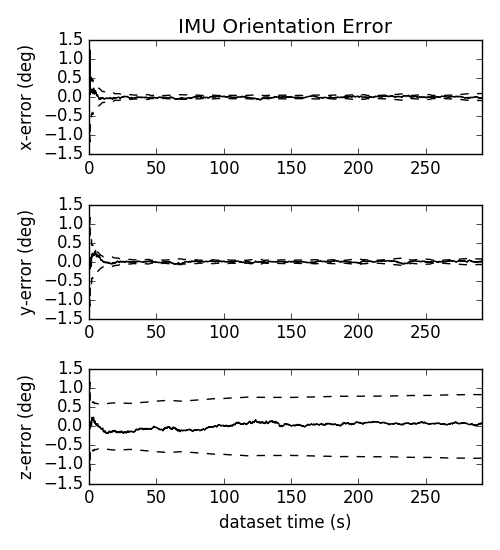}}
\subfloat{\includegraphics[width=0.32\textwidth]{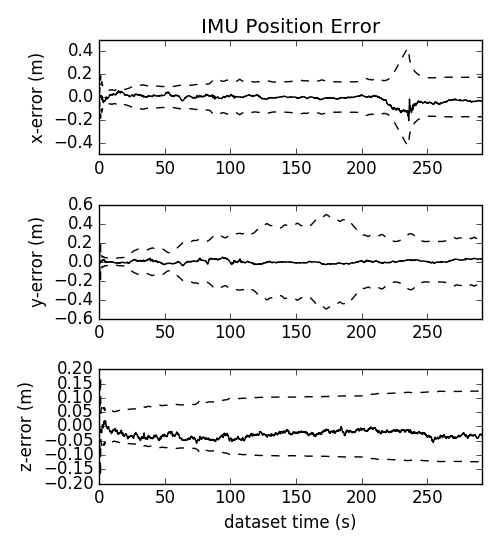}}
\subfloat{\includegraphics[width=0.32\textwidth]{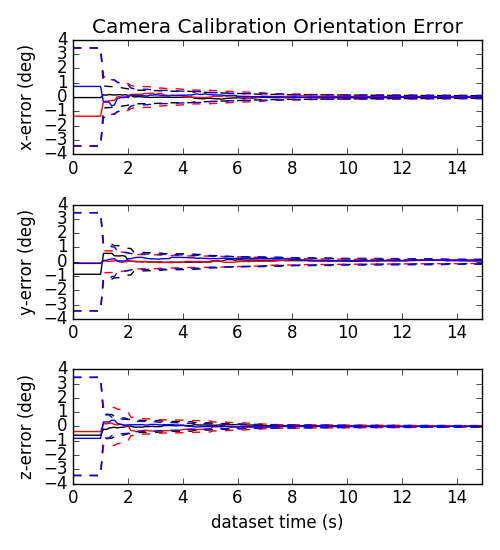}} \\[-1em]
\subfloat{\includegraphics[width=0.32\textwidth]{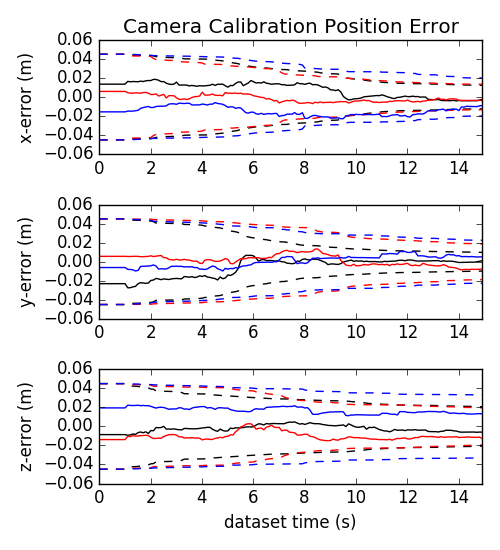}}
\subfloat{\includegraphics[width=0.32\textwidth]{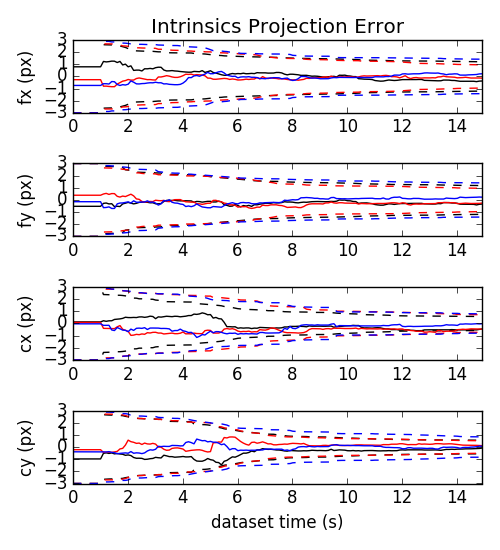}}
\subfloat{\includegraphics[width=0.32\textwidth]{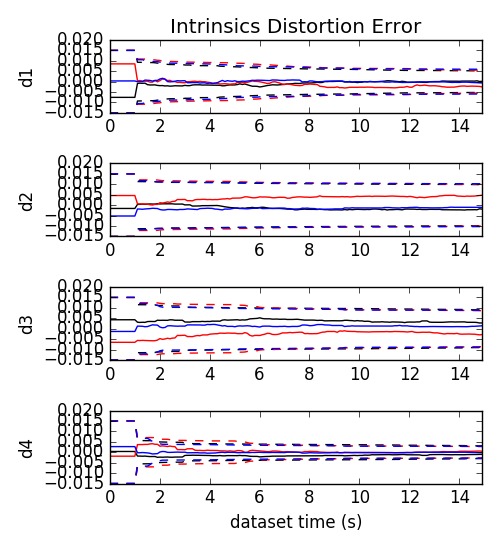}} \\[-1em]
\subfloat{\includegraphics[width=0.32\textwidth]{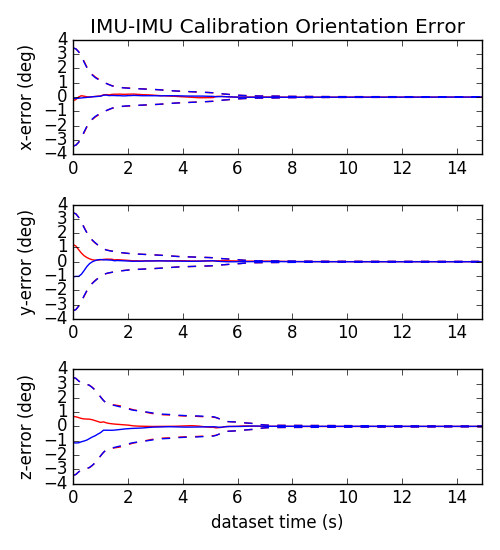}}
\subfloat{\includegraphics[width=0.32\textwidth]{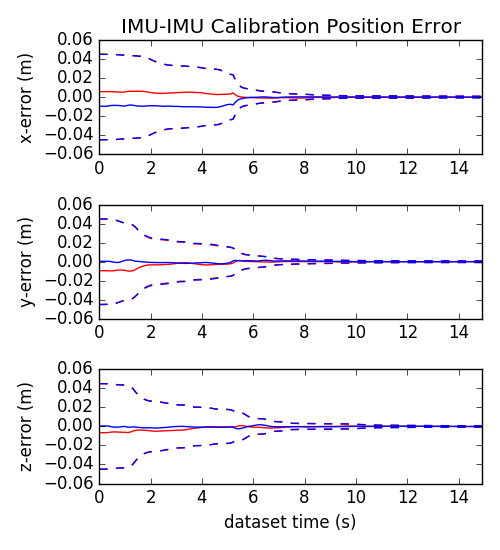}} \\[-1em]
\subfloat{\includegraphics[width=0.32\textwidth]{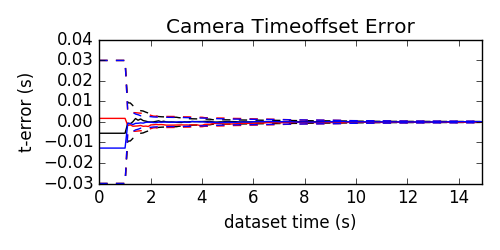}}
\subfloat{\includegraphics[width=0.32\textwidth]{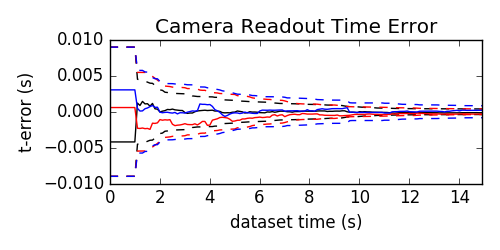}}
\subfloat{\includegraphics[width=0.32\textwidth]{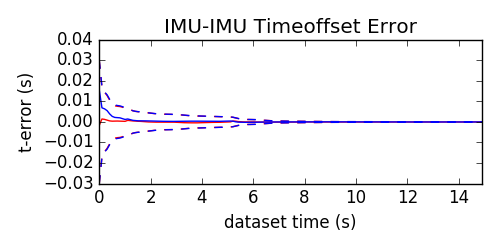}}
\caption{
The proposed MIMC-VINS calibration for a three camera and IMU configuration on the simulated Tum dataset.
Black, red and blue respectively denote parameters relating to the base, second, and third camera/IMU respectively.
For calibration parameters, only the first fifteen seconds are shown.
Solid lines refer to the errors, while dotted lines are the 3$\sigma$ reported by the estimator, demonstrating consistency of the estimator.
Note that for camera time offset the time offset of the base shown (black) is the offset between the base camera and base IMU as compared to the camera-camera time offset.
}
\label{fig:full_sim_3sigmas}
\end{figure*}

We validated the consistency of the pose and calibration estimates for a three IMU and three camera MIMC system on the synthetic Tum dataset.
We employed an interpolation order of three and the cameras were simulated at 10, 11, and 13 fps, and the RS readout times of 10, 15, and 20 milliseconds, respectively, 
while the estimator started with the initial guess of 0 for each.
As shown in Figure~\ref{fig:full_sim_3sigmas}, 
the estimation errors reveal good consistency (that is, they stay within the 3$\sigma$ bounds reported by the covariance), and additionally all calibration converges to the correct value.
For these experiments, 
we  found that inflating the noise associated with the auxiliary camera measurements during the first few seconds of the simulation leads to better consistency and convergence;
and similarly,  inflating the noise of the multi-IMU rigid-body constraints during the first few seconds of the run before switching to a very small value (rather than purely zero) offered the best performance/stability.
Thus the IMU calibration parameters can be seen to change from slow to fast convergence after this switch (in these experiments, we used the first 5 seconds).
Overall, these results validate the proposed MIMC-VINS calibration performance and consistency.

\subsubsection{FEJ Impact} \label{sec:sim_fej_compare}

We now look to investigate the impact of using FEJ within our MIMC-VINS framework.
As shown in Table \ref{table::ate_fej} and \ref{table::rpe_fej}, there is a clear advantage to using FEJ.
Even as more sensors are added the benefit of using FEJ to improve estimator consistency has a clear impact on the final accuracy of the system.
It is interesting to see that without FEJ, three camera/IMU sensors are needed to outperform the single camera/IMU pair which uses FEJ,
which further motivates us to leverage FEJ to improve estimation performance. 

\begin{table*}
\centering
\caption{ATE simulation errors in degrees/meters of the proposed MIMC-VINS with and without FEJ enabled.}
\begin{tabular}{rccccc} \toprule 
& \textbf{Num. IMU / Cam.} & \textbf{Outdoor} & \textbf{Tum} & \textbf{Gore} & \textbf{Average} \\\midrule
\multirow{3}{*}{With FEJ}
& 1, 1 & 0.239 / 0.886 & 0.237 / 0.153 & 0.768 / 0.223 & 0.415 / 0.421 \\
& 2, 2 & 0.174 / 0.588 & 0.224 / 0.115 & 0.466 / 0.153 & 0.288 / 0.285 \\
& 3, 3 & 0.145 / 0.525 & 0.177 / 0.113 & 0.307 / 0.112 & 0.210 / 0.250 \\
\midrule
\multirow{3}{*}{Without FEJ}
& 1, 1 & 0.351 / 1.048 & 0.369 / 0.194 & 2.345 / 0.428 & 1.021 / 0.557 \\
& 2, 2 & 0.257 / 0.767 & 0.325 / 0.151 & 1.165 / 0.236 & 0.583 / 0.385 \\
& 3, 3 & 0.297 / 0.859 & 0.330 / 0.151 & 0.760 / 0.173 & 0.462 / 0.394 \\
 \bottomrule
\end{tabular}
\label{table::ate_fej}
\end{table*}

\begin{table*}
\centering
\caption{RPE simulation errors in degrees/meters of the proposed MIMC-VINS with and without FEJ enabled.}
\begin{tabular}{rccccccc} \toprule 
& \textbf{Num. IMU / Cam.}  & \textbf{8m} & \textbf{16m} & \textbf{24m} & \textbf{32m} & \textbf{40m} & \textbf{48m} \\\midrule
\multirow{3}{*}{With FEJ}
& 1, 1 & 0.055 / 0.045 & 0.071 / 0.064 & 0.081 / 0.080 & 0.090 / 0.093 & 0.096 / 0.103 & 0.103 / 0.111 \\
& 2, 2 & 0.042 / 0.032 & 0.054 / 0.045 & 0.062 / 0.055 & 0.069 / 0.064 & 0.073 / 0.071 & 0.078 / 0.076 \\
& 3, 3 & 0.036 / 0.027 & 0.046 / 0.038 & 0.053 / 0.047 & 0.058 / 0.055 & 0.062 / 0.061 & 0.065 / 0.065 \\
\midrule
\multirow{3}{*}{Without FEJ}
& 1, 1 & 0.057 / 0.046 & 0.075 / 0.064 & 0.087 / 0.081 & 0.098 / 0.096 & 0.106 / 0.107 & 0.114 / 0.119 \\
& 2, 2 & 0.044 / 0.033 & 0.058 / 0.047 & 0.068 / 0.058 & 0.077 / 0.068 & 0.084 / 0.075 & 0.090 / 0.082 \\
& 3, 3 & 0.037 / 0.027 & 0.048 / 0.039 & 0.057 / 0.049 & 0.064 / 0.058 & 0.068 / 0.064 & 0.073 / 0.069 \\
\bottomrule
\end{tabular}
\label{table::rpe_fej}
\end{table*}

\subsubsection{Resilience to Sensor Failures}

\begin{figure}
\centering
\includegraphics[width=0.49\columnwidth]{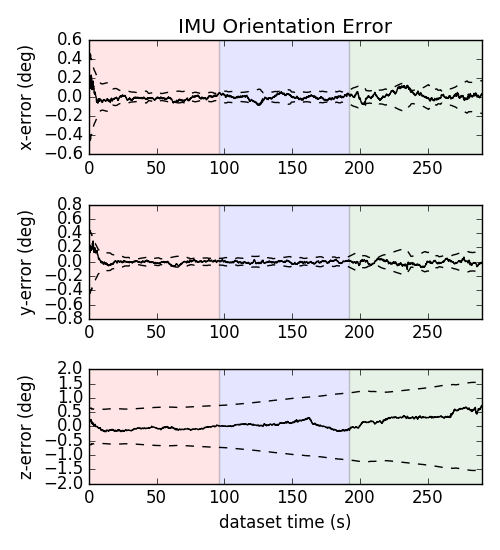} \hfill
\includegraphics[width=0.49\columnwidth]{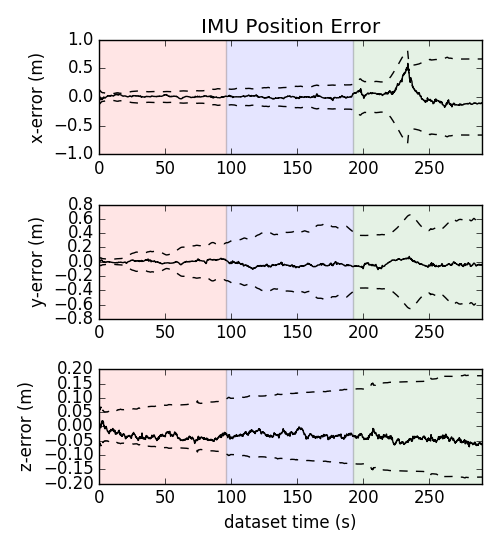}
\caption{
The orientation and position of the proposed MIMC-VINS in the presence of sensor dropouts.
The segments of the trajectory are broken into three parts: in the first (red), all three components are operational.
In the second (blue), the first component pair has failed.
In the final (green), the second component pair has also failed, leaving only one IMU-camera pair operational.
Despite these failures, redundant sensing allows for continuous and accurate estimation of the platform.
}
\label{fig:exp_results_dropout}
\end{figure}

In order to validate the proposed MIMC-VINS robustness to sensor dropouts, we simulated three cameras and three IMUs travelling along the Tum trajectory.
For this experiment, each IMU was associated with a partner camera, such that the entire system was made up of three components.
At 96 seconds, the first component that contained the base IMU and camera was turned off, such that no more camera or IMU data were generated.
This forced a switch to utilizing the second component as the base.
Note that even though no actual base camera data was collected, we still generated stochastic clones at the same rate as if it were still active.
At 192 seconds, we simulated a failure of the second component, so that only the third remained.
The pose errors and their $3\sigma$ bounds are shown in Figure~\ref{fig:exp_results_dropout}.
As evident, despite the fact that 4 out of 6 sensors have failed throughout the run, including the base IMU, 
the proposed MIMC-VINS is resilient to provide continuous and accurate pose estimates of the sensor platform.

It should be pointed out that the continuous state estimates always refer to the same sensor frame
in order for ease of external systems to continuously use these localization outputs without any interruption even in the case of sensor failure.
That is, the first-ever base IMU sensor frame, denoted $\{I_0\}$, is selected for which the state estimate is always expressed in.
To achieve this, we maintain the transformation from the frame $\{I_0\}$ to the current base IMU frame in the state even if the base IMU has failed. 
Its estimate will be updated over time and will have the same covariance propagation applied to it if additional base IMU fails.
This  estimated transformation is used to obtain the state estimate in  $\{I_0\}$ by transforming the current base IMU pose estimate into it 
and performing covariance propagation to ensure that the correct covariance of the pose estimate in the global frame is published.

 \section{Real-World Experimental Results} \label{sec:exp}

\begin{figure}
\centering
\includegraphics[width=.99\columnwidth]{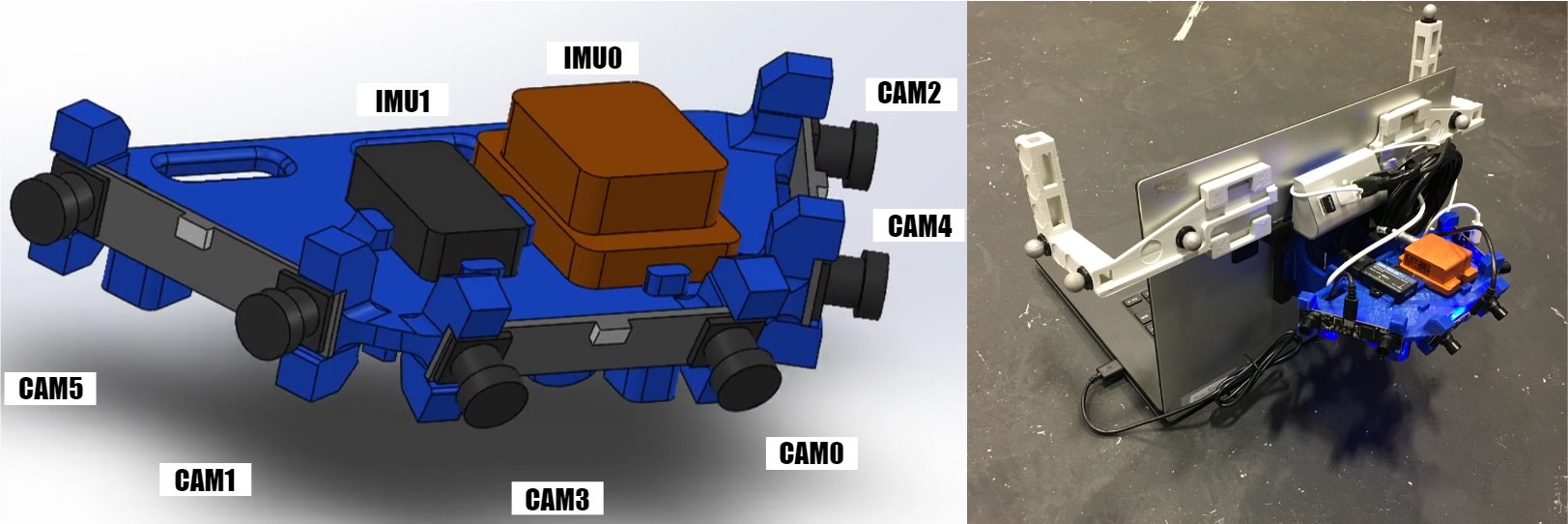}
\caption{
The sensor platform used in our experiments consists of three RS-stereo pairs (only left cameras used), XSENS MT-100, and Microstrain 3DM-GX-25 IMUs.
}
\label{fig:exp_arl_sensor_rig}
\end{figure}

To further evaluate the proposed MIMC-VINS on real-world data, 
we have built a MIMC sensor platform that consists  of three 640x480 ELP-960P2CAM-V90-VC USB 2.0 RS-stereo cameras operating at 30 Hz,
and two IMUs including an XSENS MT-100 and Microstrain 3DM-GX-25 (see Figure \ref{fig:exp_arl_sensor_rig}).
Note that the three stereo pairs were placed in a semi-circular pattern giving an overall large field of view greater then 180$^{\circ}$
while {\em only} the left image of each stereo pair was leveraged in the experimental results presented below.
Multiple datasets were collected in a large Vicon warehouse,
providing highly accurate groundtruth for comparison.
The trajectories of these datasets (labeled as multicam 1-4) are 74, 91, 185, and 108 meters in total length, respectively.

In these experiments, we initialized the estimator from a stationary position.
In particular, the base IMU collected accelerometer data during this period to estimate the direction of gravity, while the initial velocity estimate of each IMU was set to zero.
Once a sufficient excitation of the base IMU was detected, the system computes initial pose estimates of the auxiliary IMUs using the prior calibration as in the simulations. 
After initialization, propagation and update began as normal as outlined in Algorithm \ref{alg:MIMC-VINS}.
All sensing parameters were calibrated online with the initial guesses provided by offline calibration using the Kalibr toolbox \cite{Furgale2013IROS,Furgale2015IJRR,Oth2013CVPR}.
We also found that there was about 18ms readout time for the RS cameras used.

Features were extracted uniformly using FAST \cite{Rosten2010PAMI} and tracked independently (see Section \ref{sec:parallel-tracking})
for each camera's image stream using KLT \cite{Baker2004IJCV} along with outlier rejection via 8-point RANSAC.
Each camera tracked at most 100 features in its image stream.
Up to 25 features that were tracked longer than the sliding window size of 12 were added as SLAM features into the state vector.
For other features,  the MSCKF update was performed if the feature's first measurement was collected before the second oldest sliding window clone 
(i.e. if it has measurements that would become ``too old'' after the next marginalization phase).
When processing measurement updates, we slightly inflated the assumed noise sigma to 2 pixels in order to account for the unmodeled effect of image patch warping due to the RS images,
which we found experimentally led to improved accuracy of our system.\footnote{A video of these experiments is available: \url{http://www.udel.edu/007455}.}

\subsection{Estimation Accuracy Validation}

To further validate our choice of increased VI-sensors, we processed each of the datasets (multicam 1-4) using different numbers of IMUs and cameras.
To compensate for the randomness caused by RANSAC-based frontends and parallelization, we performed 30 runs for each sensor combination on each dataset.
The ATE and RPE results are given in Tables \ref{table::ate_mimc} and \ref{table::rpe_mimc},
which show that incorporating additional sensing tends to greatly improve both metrics of estimation accuracy.
In particular, the 3-camera, 2-IMU average ATE of 1.248 degrees and 0.193 meters was a great improvement over the single-camera, single-IMU which had an ATE of 2.716 degrees and 0.302 meters.
We do note, however, that this improvement was not seen in {\em all} cases such as multicam\_1,
primarily due to suboptimal tuning of the noise characteristics, especially for the IMU noises. 
In addition, we note that because the system was run in realtime, increasing the number of sensors may lead to more dropped frames in certain instances. 
This can be handled, for example, by more intelligent selection that spreads measurements \textit{across} cameras.
Importantly, adding more sensors led to improved performance in RPE at every tested segment length.
Overall, these results again confirm our desire to add additional sensors into the system,
and demonstrate that real-world accuracy gains can be achieved even without requiring synchronized or calibrated sensors.

\begin{table*}
\centering
\caption{ATE  in degrees/meters on the real-world datasets for MIMC-VINS. Average of 30 runs.}
\begin{tabular}{cccccc}\toprule
\textbf{Num. IMU / Cam.} & \textbf{multicam\_1} & \textbf{multicam\_2} & \textbf{multicam\_3} & \textbf{multicam\_4} & \textbf{Average} \\\midrule
1, 1 & 1.539 / 0.158 & 1.215 / 0.161 & 4.433 / 0.651 & 3.676 / 0.237 & 2.716 / 0.302 \\
1, 3 & 1.249 / 0.180 & 1.027 / 0.191 & 2.215 / 0.472 & 2.102 / 0.162 & 1.648 / 0.251 \\
2, 1 & \textbf{1.186} / \textbf{0.139} & 1.096 / 0.170 & 3.039 / 0.569 & 1.348 / 0.136 & 1.667 / 0.254 \\
2, 3 & 1.350 / 0.156 & \textbf{0.799} / \textbf{0.153} & \textbf{1.739} / \textbf{0.341} & \textbf{1.101} / \textbf{0.123} & \textbf{1.248} / \textbf{0.193}
 \\
\bottomrule
\label{table::ate_mimc}
\end{tabular}
\end{table*}

\begin{table*}
\centering
\caption{RPE in degrees/meters on the real-world datasets for the proposes algorithm. Average of 30 runs.}
\begin{tabular}{ccccccc} \toprule
\textbf{Num. IMU / Cam.} & \textbf{8m} & \textbf{16m} & \textbf{24m} & \textbf{32m} & \textbf{40m} & \textbf{48m} \\\midrule
1, 1 & 0.598 / 0.120 & 0.749 / 0.157 & 0.916 / 0.149 & 1.135 / 0.184 & 1.343 / 0.215 & 1.579 / 0.229 \\
1, 3 & 0.506 / 0.093 & 0.588 / 0.122 & 0.675 / 0.116 & 0.810 / 0.141 & 0.945 / 0.170 & 1.049 / 0.178 \\
2, 1 & 0.532 / 0.095 & 0.647 / 0.127 & 0.757 / 0.120 & 0.879 / 0.150 & 1.024 / 0.172 & 1.124 / 0.179 \\
2, 3 & \textbf{0.463} / \textbf{0.076} & \textbf{0.552} / \textbf{0.103} & \textbf{0.636} / \textbf{0.096} & \textbf{0.752} / \textbf{0.119} & \textbf{0.849} / \textbf{0.152} & \textbf{0.906} / \textbf{0.157} \\
\bottomrule
\label{table::rpe_mimc}
\end{tabular}
\end{table*}

\subsection{Real-Time Performance Analysis}

\begin{table*}[h]
\centering
\caption{
Timing analysis of different system components, units are in seconds.
}
\begin{tabular}{cccccccc} \toprule
\textbf{Num. IMU / Cam.} & \textbf{Track.} & \textbf{Prop.} & \textbf{Up. MSCKF} & \textbf{Up. SLAM} & \textbf{Init SLAM} & \textbf{Marg.} & \textbf{Total} \\\midrule
1, 1 & 0.0058 & 0.0007 & 0.0040 & 0.0029 & 0.0032 & 0.0013 & 0.0179 \\
1, 3 & 0.0043 & 0.0005 & 0.0127 & 0.0017 & 0.0023 & 0.0026 & 0.0240 \\ 
2, 1 & 0.0054 & 0.0015 & 0.0038 & 0.0029 & 0.0033 & 0.0016 & 0.0184 \\
2, 3 & 0.0046 & 0.0011 & 0.0129 & 0.0017 & 0.0029 & 0.0034 & 0.0266 \\ 
\bottomrule
\end{tabular}
\label{table::timing_components}
\end{table*}

\begin{figure}
\centering
\includegraphics[width=0.99\columnwidth]{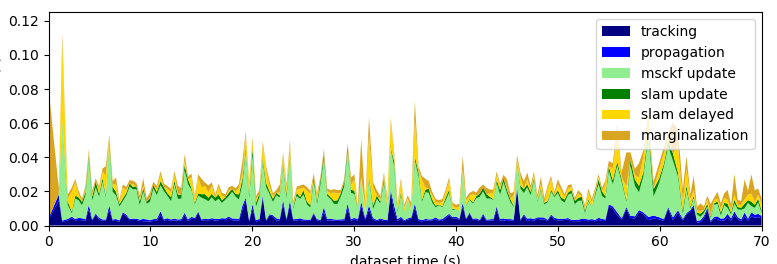}
\includegraphics[width=0.99\columnwidth]{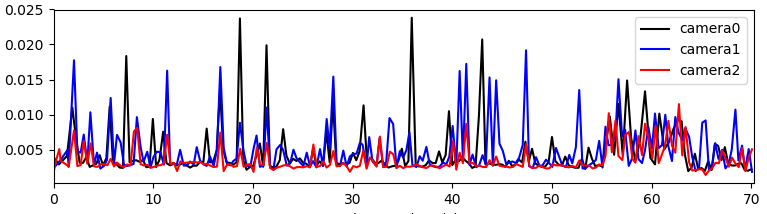}
\caption{
Example timing results of the 3 camera 2 IMU MIMC-VINS configuration on multicam\_1.
A per-component timing (top) of the base camera tracking and the state update can be seen.
Additionally, the timing of each camera's tracking thread is shown (bottom). While Table~\ref{table::timing_components} illustrates that the average frame processing times remain below the the 30 Hz rate of the cameras, there exists spikes in computation which force the dropping of frames. This can be mitigated by more resource-aware selection of features for processing.
}
\label{fig:timing_plots}
\end{figure}

To evaluate the realtime performance of the proposed MIMC-VINS estimator, 
we have timed both the complete update thread corresponding to the base camera along with each asynchronous tracking thread for each additional camera.
The timing results for the 3-camera 2-IMU configuration are shown in Figure \ref{fig:timing_plots}, 
from which we can see that the individual components typically perform below 0.040 seconds of total execution time, while on average the it takes 0.0266 seconds to perform an update.
Note that the system for all the experiments was evaluated on an Intel Core  i7-7700HQ CPU clocked at 2.80GHz base frequency.

In the case where the update takes more than the sensing rate, in these datasets this would be 30 Hz, the next frame is dropped to ensure realtime pose results.
The majority of the cost comes from the MSCKF feature update which is expected as propagation typically plays a very small contribution in execution time, and the visual tracking has been parallelized.
For the per-thread tracking time for each camera,
it is clear that all cameras take about the same time to track (although with occassional spikes), as each camera has the same resolution and number of feature tracks, 
and the parallelization works as expected to save computation time on non-base feature measurements.
A more detailed breakdown of the average processing time is shown in Table \ref{table::timing_components} of all different sensor configurations. The results show the average timing for feature tracking, propagation (including multi-IMU update), MSCKF update, SLAM update, new SLAM feature initialization, and marginalization of old states.
Since the addition of each camera adds more feature measurements that are included in each update, we expected to see an increase in the update time.
While propagation is expected to also increase, its additional computational cost is negligible as compared to the cost of additional feature measurements which are available for update.
In the future we will investigate optimal selection of measurements for update to bound the update computational cost even as the number of features tracked are added.

\subsection{Sensor Resilience Evaluation}

\begin{figure}
\centering
\includegraphics[width=0.8\columnwidth]{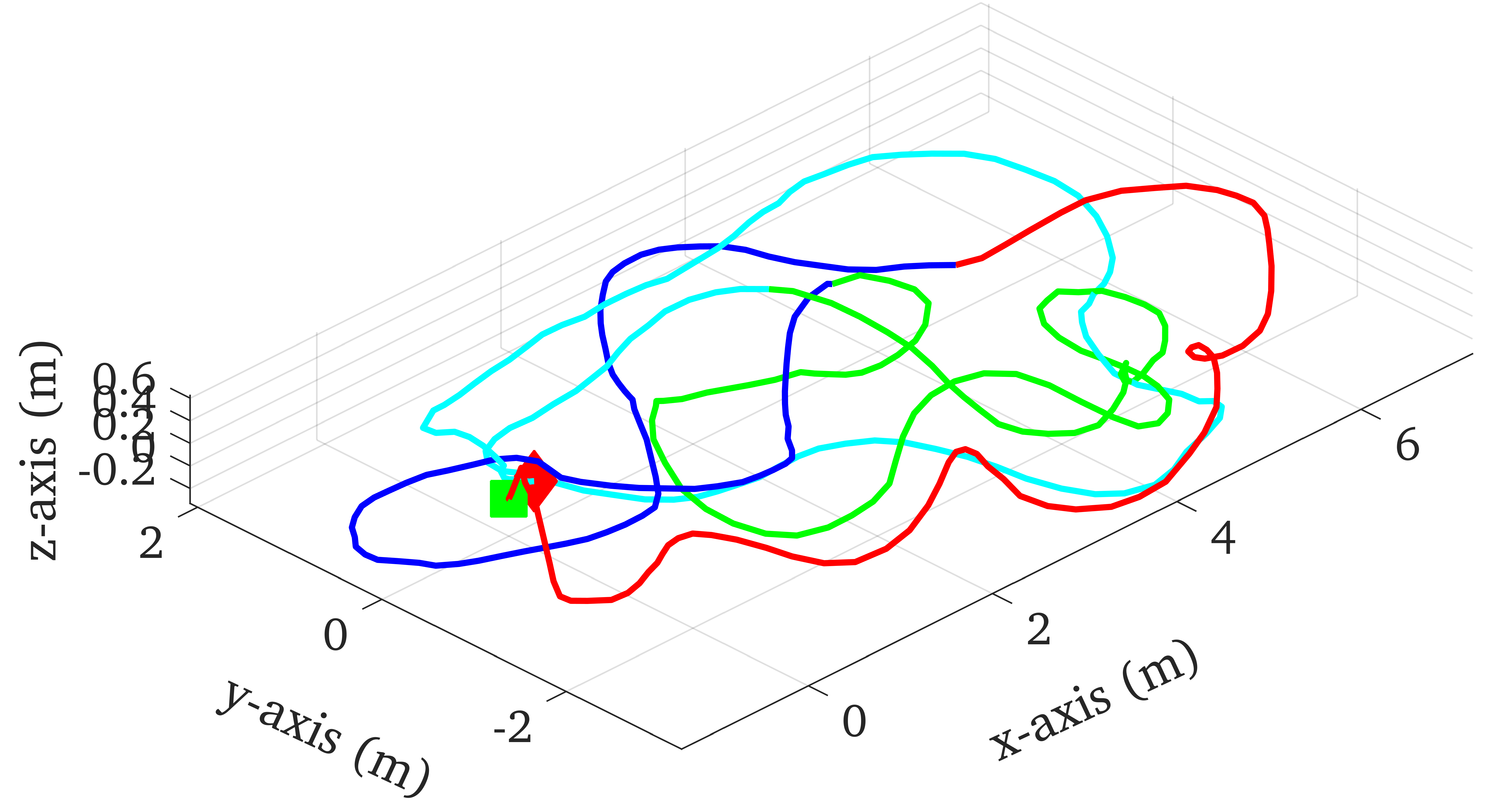}
\caption{
The trajectory estimates of the proposed MIMC-VINS on multicam\_1 dataset in the presence of sensor dropouts.
Red section has 3 camera and 2 IMU, blue has 2 camera and 2 IMU, green has 2 camera and 1 IMU, and cyan has 1 camera and 1 IMU.
}
\label{fig:exp_arl_dropout_traj}
\end{figure}

\begin{figure}
\centering
\includegraphics[width=0.49\columnwidth]{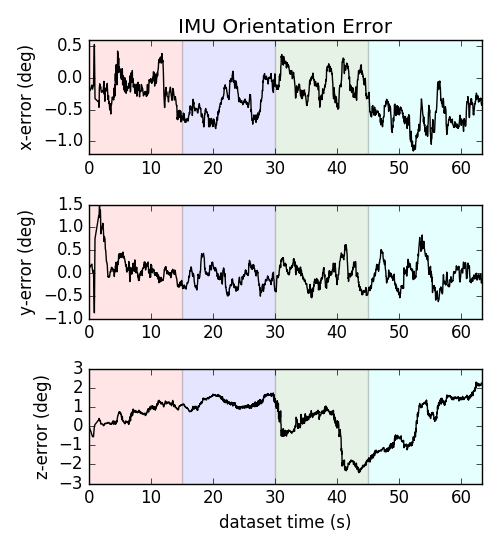} \hfill
\includegraphics[width=0.49\columnwidth]{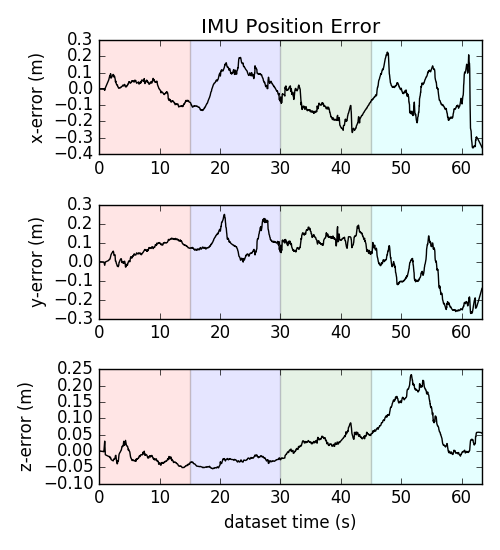}
\caption{
The pose estimation errors of the proposed MIMC-VINS on multicam\_1 dataset in the presence of sensor dropouts.
The first camera is dropped out at 15 seconds, the first IMU at 30 seconds, and the second camera at 45 seconds into the dataset.
}
\label{fig:exp_arl_dropout}
\end{figure}

To validate the resilience of the proposed MIMC-VINS  to sensor failure in the real world, 
we demonstrated a series of sequential failures on the multicam\_1 dataset
and  the trajectory estimate and RMSE results after alignment to the groundtruth are shown 
in Figure \ref{fig:exp_arl_dropout_traj} and  \ref{fig:exp_arl_dropout}.
In particular, at the beginning of the dataset the system has all three cameras and two IMUs.
The cameras are then failed sequentially at 15 and 45 seconds into the dataset, 
while for the IMUs there is only a single failure of the \textit{base} in the middle of the dataset at 30 seconds.
The error and trajectory estimates are continuous and thus, there are not large ``jumps'' in the error at these failure points.
This validates the proposed method's robustness to sensor failures and its ability to provide continuous state estimates.
It is also important to note that as more sensors are being lost the accuracy is expected to decrease (see Section \ref{sec:sim}), 
which explains towards the end of the dataset the position error starts to increase at a faster rate due to only having information from a single camera.

 \section{Conclusions and Future Work}
\label{sec:concl}

In this paper we have developed a versatile and resilient  multi-IMU multi-camera (MIMC)-VINS  algorithm 
that is able to seamlessly fuse the multi-modal visual-inertial information from an arbitrary number of uncalibrated cameras and IMUs.
Within an efficient and consistent multi-state constraints Kalman filter (MSCKF) framework,
the proposed MIMC-VINS estimator is able to preserve the similar computational complexity as in the single-IMU/camera case 
(i.e. only moderately increasing the size of the state vector if more sensors are used),
while providing smooth, uninterrupted, accurate 3D motion tracking even if some sensors fail.
We have extensively validated the proposed approach in both Monte-Carlo simulations and real-world experiments.
In the future, we will investigate how to make the proposed MIMC-VINS adaptive to (computational) resource constraints,
and will also extend this work to cooperative visual-inertial estimation for distributed multi-robot systems.

\begin{appendices}
\section{Interpolation Measurement Jacobians}
 \label{appendix::HOI}

To simplify derivations, we define the following matrices:
\begin{align}
    \mathbf{m}(t) &= \begin{bmatrix} \Delta t\mathbf{I}_3 & 
    \Delta t^2\mathbf{I}_3 & \cdots & \Delta t^n\mathbf{I}_3
    \end{bmatrix} \\
    \mathbf{A}_\theta &= \mathbf{m}(t)\mathbf{a}_\theta \\
    \mathbf{A}_p &= \mathbf{m}(t)\mathbf{a}_p
\end{align}
Lastly, we also define the left Jacobian of SO(3), $\mathbf{J}_l\left(\cdot\right)$~\cite{chirikjian2011stochastic}.
Then we can write the orientation interpolation as:
\begin{align}
    {}^{I(t)}_G{\mathbf{R}} &= \mathbf{Exp}\left(\mathbf{A}_\theta\right){}^{I(t_0)}_G{\mathbf{R}}
\end{align}
We wish to find the Jacobian of this pose in respect to the poses the polynomial. We next perturb each side:
\begin{align}
&\textrm{Exp}\left(-{}^{I(t)}_G\tilde{\bm \theta}\right){}^{I(t)}_G\hat{\mathbf{R}} \notag \\ &= 
\textrm{Exp}\left(\hat{\mathbf{A}}_\theta+\tilde{\mathbf{A}}_\theta\right)\textrm{Exp}\left(-{}^{I(t_0)}_G\tilde{\bm \theta}\right){}^{I(t_0)}_G\hat{\mathbf{R}} \notag \\
&\approx \textrm{Exp}\left(\mathbf{J}_l\left(\hat{\mathbf{A}}_\theta\right)\tilde{\mathbf{A}}_\theta\right)\textrm{Exp}\left(\hat{\mathbf{A}}_\theta\right)\textrm{Exp}\left(-{}^{I(t_0)}_G\tilde{\bm \theta}\right){}^{I(t_0)}_G\hat{\mathbf{R}} \notag \\
&\scalemath{.9}{= \textrm{Exp}\left(\mathbf{J}_l\left(\hat{\mathbf{A}}_\theta\right)\tilde{\mathbf{A}}_\theta\right)\textrm{Exp}\left(-\textrm{Exp}\left(\hat{\mathbf{A}}_\theta\right){}^{I(t_0)}_G\tilde{\bm \theta}\right)\textrm{Exp}\left(\hat{\mathbf{A}}_\theta\right){}^{I(t_0)}_G\hat{\mathbf{R}}} \notag\\
& \approx 
\textrm{Exp}\left(\mathbf{J}_l\left(\hat{\mathbf{A}}_\theta\right)\tilde{\mathbf{A}}_\theta-\textrm{Exp}\left(\hat{\mathbf{A}}_\theta\right){}^{I(t_0)}_G\tilde{\bm \theta}\right){}^{I(t)}_G\hat{\mathbf{R}}
\end{align}
Thus we can immediately pull out Jacobians as:
\begin{align}
    \frac{\partial {}^{I(t)}_G\tilde{\bm \theta}}{\partial 
    {}^{I(t_0)}_G\tilde{\bm \theta}} &= -\mathbf{J}_l\left(\hat{\mathbf{A}}_\theta\right) \frac{\partial \tilde{\mathbf{A}}_\theta}{\partial 
    {}^{I(t_0)}_G\tilde{\bm \theta}} + \textrm{Exp}\left(\hat{\mathbf{A}}_\theta\right) \\
    \frac{\partial {}^{I(t)}_G\tilde{\bm \theta}}{\partial 
    {}^{I(t_k)}_G\tilde{\bm \theta}} &= -\mathbf{J}_l\left(\hat{\mathbf{A}}_\theta\right) \frac{\partial \tilde{\mathbf{A}}_\theta}{\partial 
    {}^{I(t_k)}_G\tilde{\bm \theta}} 
\end{align}
In order to compute this, we also perturb $\mathbf{A}_\theta$:
\begin{align}
    \hat{\mathbf{A}}_\theta + \tilde{\mathbf{A}}_\theta &= \mathbf{m}(t)\left(\hat{\mathbf{a}}_\theta + \tilde{\mathbf{a}}_\theta \right) \\
    &= \mathbf{m}(t)\left(\hat{\mathbf{a}}_\theta + \mathbf{V}^{-1}\Delta \tilde{\bm \phi} \right) \\
    \Rightarrow
    \frac{\partial \tilde{\mathbf{A}}_\theta}{\partial 
    {}^{I(t_i)}_G\tilde{\bm \theta}} &= \mathbf{m}(t)\mathbf{V}^{-1}\frac{\partial \Delta \tilde{\bm \phi}}{\partial 
    {}^{I(t_i)}_G\tilde{\bm \theta}}
\end{align} \\
Next, we perturb $\Delta {\bm \phi}$:
\begin{align}
    \Delta \hat{\bm \phi} + \Delta \tilde{\bm \phi} &= \begin{bmatrix}
   \Delta \hat{\bm \phi}_1 + \Delta \tilde{\bm \phi}_1 \\
    \Delta \hat{\bm \phi}_2 + \Delta \tilde{\bm \phi}_2 \\
    \vdots \\
    \Delta \hat{\bm \phi}_n + \Delta \tilde{\bm \phi}_n
    \end{bmatrix} \\
    \Delta \hat{\bm \phi}_k + \Delta \tilde{\bm \phi}_k &= 
    \textrm{Log}\left(\textrm{Exp}\left(-{}^{I(t_k)}_G \tilde{\bm \theta} \right){}^{I(t_k)}_G\hat{\mathbf{R}} {}^{G}_{I(t_0)}\hat{\mathbf{R}} \right) \notag\\
    &\approx \textrm{Log}\left(\textrm{Exp}\left(\Delta \hat{\bm \phi}_k-\mathbf{J}_l^{-1}\left(\Delta \hat{\bm \phi}_k\right){}^{I(t_k)}_G \tilde{\bm \theta} \right)\right) \notag\\
    &= \Delta \hat{\bm \phi}_k-\mathbf{J}_l^{-1}\left(\Delta \hat{\bm \phi}_k\right){}^{I(t_k)}_G \tilde{\bm \theta} \\
     \Delta \hat{\bm \phi}_k + \Delta \tilde{\bm \phi}_k &= \textrm{Log}\left({}^{I(t_k)}_G\hat{\mathbf{R}} {}^{G}_{I(t_0)}\hat{\mathbf{R}}\textrm{Exp}\left({}^{I(t_0)}_G \tilde{\bm \theta} \right) \right) \notag\\
     &= \textrm{Log}\left(\textrm{Exp}\left({}^{I(t_k)}_{I(t_0)}\hat{\mathbf{R}}{}^{I(t_0)}_G \tilde{\bm \theta} \right) {}^{I(t_k)}_{I(t_0)}\hat{\mathbf{R}} \right) \notag\\
     &\scalemath{.9}{\approx \textrm{Log}\left(\textrm{Exp}\left(\Delta \hat{\bm \phi}_k + \mathbf{J}_l^{-1}\left(\Delta \hat{\bm \phi}_k\right){}^{I(t_k)}_{I(t_0)}\hat{\mathbf{R}}{}^{I(t_0)}_G \tilde{\bm \theta} \right) \right)} \notag \\
     &\Delta \hat{\bm \phi}_k + \mathbf{J}_l^{-1}\left(\Delta \hat{\bm \phi}_k \right){}^{I(t_k)}_{I(t_0)}\hat{\mathbf{R}}{}^{I(t_0)}_G \tilde{\bm \theta}
\end{align}
Thus we have:
\begin{align}
    \frac{\partial \Delta \tilde{\bm \phi}}{\partial {}^{I(t_0)}_G \tilde{\bm \theta}} &= \begin{bmatrix}
    -\mathbf{J}_l^{-1}\left(\Delta \hat{\bm \phi}_1 \right) \\
    -\mathbf{J}_l^{-1}\left(\Delta \hat{\bm \phi}_2 \right) \\
    \vdots \\
    -\mathbf{J}_l^{-1}\left(\Delta \hat{\bm \phi}_n \right)
    \end{bmatrix} \\
    \frac{\partial \tilde{\bm \theta}}{\partial {}^{I(t_k)}_G \tilde{\bm \theta}} &= \begin{bmatrix}
    \mathbf{0} \\
    \vdots \\
    \mathbf{J}_l^{-1}\left(\Delta \hat{\bm \phi}_k \right){}^{I(t_k)}_{I(t_0)}\hat{\mathbf{R}} \\
    \vdots \\
    \mathbf{0}
    \end{bmatrix}
\end{align}
In the case that we are estimating a time offset, that is ${t} = {t}_m + {}^{C_i}t_{C_b}$, then we will additionally have:
\begin{align}
    \frac{\partial {}^{I(t)}_G\tilde{\bm \theta}}{\partial 
    {}^{C_i}\tilde{t}_{C_b}} &= -\mathbf{J}_l\left(\hat{\mathbf{A}}_\theta\right) \frac{\partial \tilde{\mathbf{A}}_\theta}{\partial 
    {}^{C_i}\tilde{t}_{C_b}} \\
    \frac{\partial \tilde{\mathbf{A}}_\theta}{\partial 
    {}^{C_i}\tilde{t}_{C_b}} &= \frac{\partial \mathbf{m}(t)}{\partial {}^{C_i}\tilde{t}_{C_b}} \hat{\mathbf{a}}_\theta \\
    \frac{\partial \mathbf{m}(t)}{\partial {}^{C_i}\tilde{t}_{C_b}} &= \begin{bmatrix} \mathbf{I}_3 & 2\Delta \hat{t} \mathbf{I}_3 & \cdots  & n \Delta \hat{t}^{n-1} \mathbf{I}_3 \end{bmatrix}
\end{align} 
Next we look at the position interpolation:
\begin{align}
    \frac{\partial {}^G\tilde{\mathbf{p}}_{I(t)}}{
    \partial {}^G\tilde{\mathbf{p}}_{I(t_0)}} &= \mathbf{I}_3 + \mathbf{m}(t)\mathbf{V}^{-1}\frac{\partial \Delta \mathbf{p}}{\partial {}^G\tilde{\mathbf{p}}_{I(t_0)}} \\
    \frac{\partial \Delta \mathbf{p}}{\partial {}^G\tilde{\mathbf{p}}_{I(t_0)}} &= \begin{bmatrix}
    -\mathbf{I}_3 &
    -\mathbf{I}_3 &
    \cdots &
    -\mathbf{I}_3
    \end{bmatrix}^\top \\
    \frac{\partial {}^G\tilde{\mathbf{p}}_{I(t)}}{
    \partial {}^G\tilde{\mathbf{p}}_{I(t_k)}} &= \mathbf{m}(t)\mathbf{V}^{-1}\frac{\partial \Delta \mathbf{p}}{\partial {}^G\tilde{\mathbf{p}}_{I(t_k)}} \\
    \frac{\partial \Delta \mathbf{p}}{\partial {}^G\tilde{\mathbf{p}}_{I(t_k)}} &= \begin{bmatrix}
    \mathbf{0}_3 &
    \cdots &
    \mathbf{I}_3 &
    \cdots &
    \mathbf{0}_3 
    \end{bmatrix}^\top \\
    \frac{\partial {}^G\tilde{\mathbf{p}}_{I(t)}}{
    \partial {}^{C_i}\tilde{t}_{C_b}} &= \frac{\partial \mathbf{m}(t)}{\partial {}^{C_i}\tilde{t}_{C_b}} \hat{\mathbf{a}}_p
\end{align}

 \end{appendices}

{
\renewcommand*{\bibfont}{\footnotesize}
\printbibliography
}

\begin{IEEEbiography}[{\includegraphics[width=1in,height=1.25in,clip,keepaspectratio]{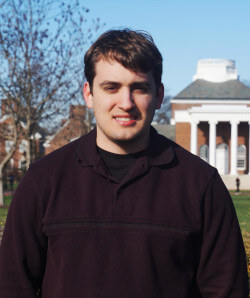}}]
{Kevin Eckenhoff} received a Bachelors in Mechanical Engineering with minors in Mathematics and Physics from the University of Delaware, USA, in 2014. He received a Ph.D. in Mechanical Engineering from the University of Delaware in 2020 under the advisement of Guoquan Huang. His research focused on visual-inertial navigation, online sensor calibration, and target tracking.

He was the recipient of the Helwig Fellowship starting in 2014. Dr. Eckenhoff currently works as a Research Scientist for Facebook. 

\end{IEEEbiography}

\begin{IEEEbiography}[{\includegraphics[width=1in,height=1.25in,clip,keepaspectratio]{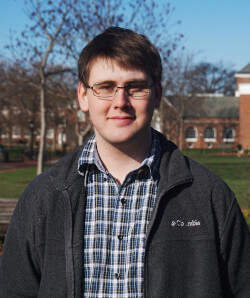}}]{Patrick Geneva}
received a Bachelors in Mechanical Engineering along with minors in Computer Science and Mathematics from University of Delaware, USA, in 2017. 

He is currently a Ph.D. in Computer Science candidate under Guoquan Huang in the Robot Perception and Navigation Group (RPNG) at the University of Delaware.
His primary research interests are on resource constrained visual-inertial state estimation, including multi-sensor systems, autonomous vehicles, and probabilistic sensor fusion.
He has received the Mary and George Nowinski Award for Excellence in Undergraduate Research in 2017 for his undergraduate research work under Dr. Huang, and the NASA Delaware Space Grant (DESG) Graduate Fellowship in 2019.

\end{IEEEbiography}

\begin{IEEEbiography}[{\includegraphics[width=1in,height=1.25in,clip,keepaspectratio]{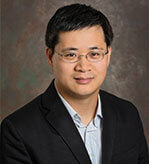}}]{Guoquan Huang}
received the B.Eng. in automation (electrical engineering) from University of Science and Technology Beijing,  China,  in  2002,  and  the  M.Sc.  and  Ph.D. in computer science from University of Minnesota--Twin Cities, Minneapolis, MN, USA, in 2009 and 2012, respectively. 

He is currently an Assistant Professor of Mechanical Engineering at the University of Delaware (UD), Newark, DE, USA, where he is leading the Robot Perception and Navigation Group (RPNG). He also holds an Adjunct Professor position at the Zhejiang University, Hangzhou, China. He was a Senior Consultant (2016-2018) at the Huawei 2012 Laboratories and a Postdoctoral Associate (2012-2014) at the MIT Computer Science and Artificial Intelligence Laboratory (CSAIL), Cambridge, MA. 
His research interests focus on state estimation and spatial AI for robotics, including probabilistic sensing, localization, mapping, perception and navigation of autonomous vehicles. 

Dr. Huang has received various honors and awards including the 2006 Academic Excellence Fellowship from the University of Minnesota, 2011 Chinese Government Award for Outstanding Self-Financed Students Abroad, 2015 UD Research Award (UDRF), 2016 NSF CRII Award, 2017 UD Makerspace Faculty Fellow, 2018 SATEC Robotics Delegation (invited by ASME), 2018 Google Daydream Faculty Research Award, 2019 Google AR/VR Faculty Research Award, and was the Finalist for the 2009 Best Paper Award from the Robotics: Science and Systems Conference (RSS).

\end{IEEEbiography}

\end{document}